\documentclass[11pt,dvipsnames,table]{article}

\newif\ifreview
\reviewfalse

\newif\ifpreprint
\preprintfalse

\ifreview
\usepackage[review]{acl}
\else
\ifpreprint
\usepackage[preprint]{acl}
\else
\usepackage[final]{acl}
\fi
\fi

\usepackage{times}
\usepackage{latexsym}

\usepackage[T1]{fontenc}

\usepackage[utf8]{inputenc}

\usepackage{microtype}

\usepackage{inconsolata}

\usepackage{graphicx}

\usepackage{xcolor}
\usepackage{subcaption}
\usepackage{wrapfig}
\usepackage{booktabs}
\usepackage{amsfonts}
\usepackage{multirow}
\usepackage{longtable}
\usepackage{pifont}
\usepackage{lipsum}
\usepackage{mdframed}
\usepackage{soul}  %
\usepackage{tablefootnote}
\usepackage{listings}
\usepackage{tcolorbox}
\usepackage{setspace}
\usepackage{amsmath}

\usepackage{tikz}
\usetikzlibrary{patterns}

\usepackage[frozencache,cachedir=.]{minted}
\definecolor{minted@bgcolor}{rgb}{0.9, 0.8, 1.0}

\usepackage{array}
\newcolumntype{M}[1]{>{\centering\arraybackslash}m{#1}}
\newcolumntype{L}[1]{>{\raggedright\arraybackslash}m{#1}}
\newcolumntype{R}[1]{>{\raggedleft\arraybackslash}m{#1}}

\usepackage{enumitem}
\setlist[itemize]{leftmargin=*,itemsep=2pt} %

\newcommand{\cmark}{\textcolor{teal}{\ding{51}}}
\newcommand{\xmark}{\textcolor{red}{\ding{55}}}

\renewcommand{\paragraph}[1]{\vskip .3em \noindent {\bf #1}}

\newtcolorbox{takeawaybox}[1]{
    colback=gray!10,
    colframe=blue!20,
    arc=4pt,
    left=2pt,
    right=2pt,
    top=2pt,
    bottom=2pt,
    boxsep=2pt,
    title={#1},
    coltitle=black,
    fonttitle=\bfseries,
    center title,
    toptitle=1pt
}

\newcommand\fover{\textsc{FoVer}}

\title{Efficient PRM Training Data Synthesis via Formal Verification}

\newcommand{\psu}{\textsuperscript{1}}
\newcommand{\cu}{\textsuperscript{2}}

\author{Ryo Kamoi\psu \quad Yusen Zhang\cu \quad Nan Zhang\psu \quad Sarkar Snigdha Sarathi Das\psu \\
{\bf Ranran Haoran Zhang\psu \quad Wenpeng Yin\psu \quad Rui Zhang\psu} \\
  \psu Penn State University \quad \cu Columbia University \\
  \texttt{\{ryokamoi, rmz5227\}@psu.edu}
\\}

\begin{document}

\maketitle

\begin{abstract}
Process Reward Models~(PRMs) have emerged as a promising approach for improving LLM reasoning capabilities by providing process supervision over reasoning traces. However, existing approaches for constructing PRM training data remain costly and noisy, as they typically rely on human annotation or sampling-based labeling methods that require repeated LLM calls. In this work, we propose \fover, a framework that synthesizes PRM training data from formal reasoning tasks by annotating step-level error labels using formal verification tools such as Z3 and Isabelle. By leveraging formal verification, \fover{} enables efficient and accurate PRM data construction without requiring human annotation or additional LLM calls. Using \fover, we create PRM training data from formal logic and theorem proving tasks. Experiments on 12~reasoning benchmarks show that fine-tuning on our training data improves PRMs not only on math and logic reasoning tasks, which are informal variants of the training tasks, but also on NLI and BBH benchmarks, which differ substantially from the tasks used to construct the training data. These results demonstrate the practical effectiveness of \fover, showing that PRM training data created using formal verification improves PRMs on informal reasoning tasks written in natural language.
\ifreview
\else
The datasets, models, and code are provided at~\url{https://github.com/psunlpgroup/FoVer}.
\fi
\end{abstract}

\let\oldaddcontentsline\addcontentsline
\renewcommand{\addcontentsline}[3]{}

\begin{figure*}[t]
    \centering
    \includegraphics[width=\linewidth]{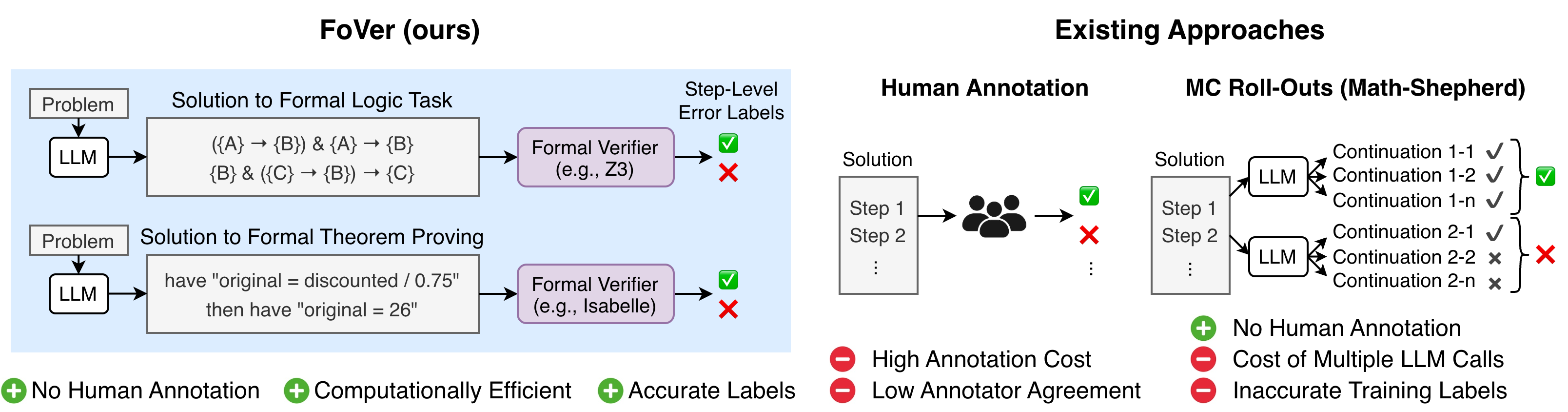}
    \caption{
    Comparison of \fover{} with existing methods for PRM training data creation. \fover{} efficiently produces accurate PRM training data for formal reasoning tasks by leveraging formal verification tools. In contrast, existing methods are costly and produce noisy labels, as they rely on human annotation or repeated LLM calls.
    }
    \label{fig:overview}
\end{figure*}

\section{Introduction}

Process Reward Models~(PRMs) have recently emerged as a promising approach for improving the reasoning capabilities of LLMs. They provide fine-grained process supervision over intermediate reasoning steps during training and inference~\citep{uesato2022solving, lightman2024lets}. In practice, PRMs are usually created by fine-tuning LLMs to classify the correctness of individual reasoning steps, using training datasets annotated with step-level error labels on LLM-generated reasoning traces~\citep{wang-etal-2024-math-shepherd, zhang2025lessons}.

\begin{figure}[t]
    \centering
    \begin{subfigure}{\linewidth}
        \includegraphics[width=\linewidth,trim={5pt 7pt 5pt 7pt},clip]{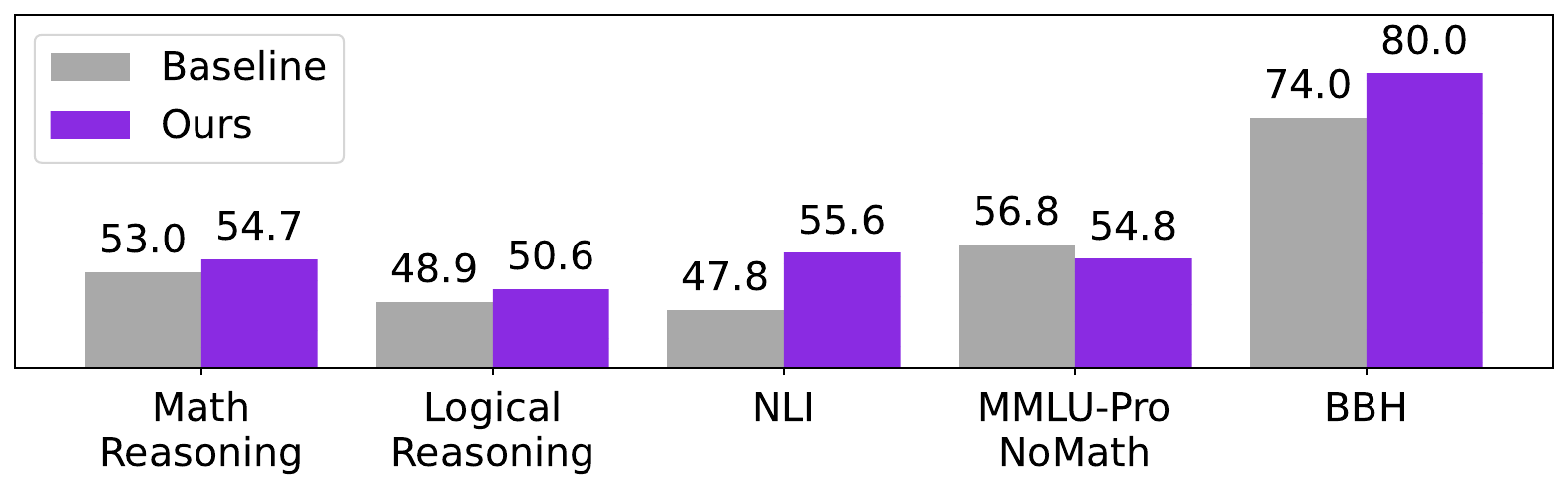}
        \caption{PRMs based on Llama 3.1 8B}
    \end{subfigure}
    \vskip 1em
    \begin{subfigure}{\linewidth}
        \includegraphics[width=\linewidth,trim={5pt 7pt 5pt 7pt},clip]{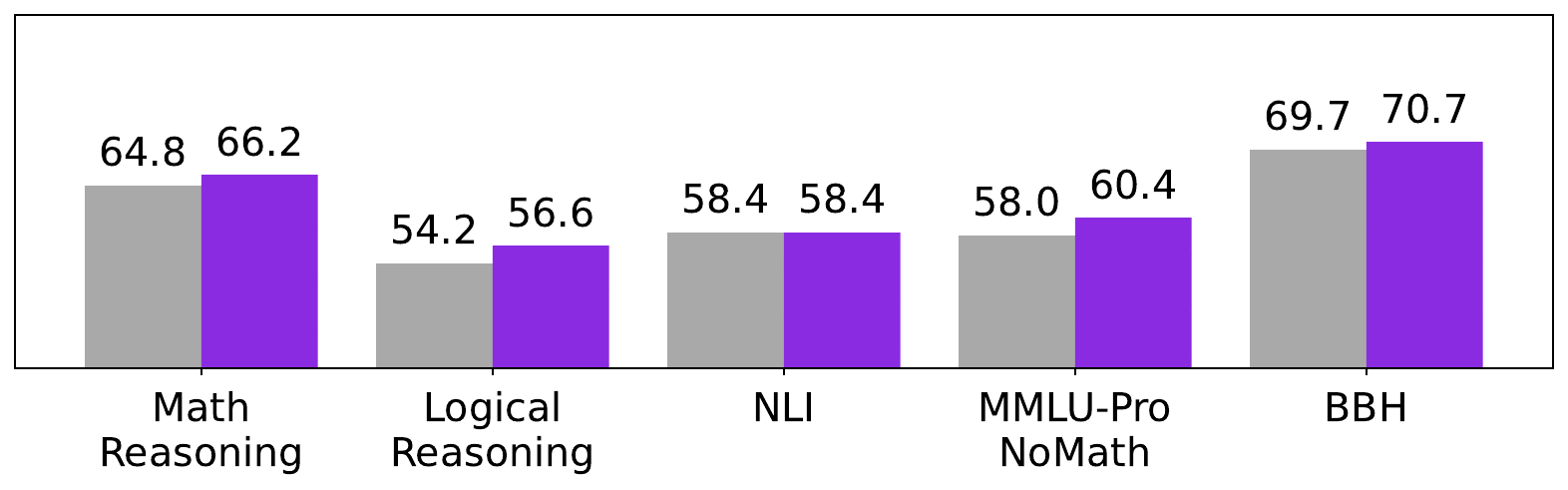}
        \caption{PRMs based on Qwen 2.5 7B}
    \end{subfigure}
    \caption{
    Best-of-7 results across 5~categories of 12~reasoning benchmarks. PRMs trained with \fover{} outperform baseline PRMs on average, showing that our training data created from formal reasoning improves PRMs on widely used reasoning benchmarks. Details are in Section~\ref{sec:experiments}.
    }
    \label{fig:performance-summary}
\end{figure}

Despite their growing adoption, creating PRM training data remains costly and noisy. Early studies~\citep{uesato2022solving, lightman2024lets} create PRM training data via human annotation, which is particularly expensive for obtaining step-level error labels and suffers from low inter-annotator agreement~\citep{zheng2024processbench}. Although Monte Carlo roll-outs~\citep{wang-etal-2024-math-shepherd, luo2024improve}, also known as Math-Shepherd, provide an annotation-free alternative for training data creation, it requires multiple LLM calls to label each step and generates noisy training labels.

To address this gap, we propose \fover, a novel framework for efficiently creating accurate PRM training data. 
As shown in Figure~\ref{fig:overview}, \fover{} leverages formal verification tools to assign accurate step-level error labels to reasoning traces for formal reasoning tasks such as logic and theorem proving, without relying on human annotation or repeated LLM calls.
Using \fover, we create PRM training data named \fover-40K by annotating step-level error labels on LLM responses to formal logic and theorem proving tasks using Z3~\citep{demoura2008z3} and Isabelle~\citep{nipkow2002isabelle}. We then fine-tune Llama 3.1 8B~\citep{dubey2024llama3} and Qwen~2.5~7B~\citep{qwen25} on \fover-40K to use them as PRMs, which we refer to as \fover-PRMs.

Although \fover{} creates PRM training data from formal reasoning tasks, PRMs are typically used to detect mistakes in informal reasoning tasks written in natural language. Therefore, in our experiments, we evaluate PRMs trained with \fover{} on widely used informal reasoning benchmarks.
First, we evaluate \fover-PRMs on informal logic and mathematical reasoning benchmarks. These benchmarks can be regarded as informal variants of the training tasks, as \fover-40K consists of formal logic and theorem proving tasks. We conduct evaluations on 6~widely used benchmarks, including LogicNLI~\citep{tian-etal-2021-diagnosing}, AQuA-RAT~\citep{ling-etal-2017-program}, and AIME~\citep{aops_aime}. Using Best-of-K~\citep{cobbe2021gsm8k, li-etal-2023-making}, a standard evaluation approach for PRMs~\citep{lightman2024lets, wang-etal-2024-math-shepherd}, we show that \fover-PRMs outperform baseline PRMs based on LLMs without additional fine-tuning.
Second, we evaluate \fover-PRMs on informal reasoning benchmarks involving tasks that substantially differ from those used to construct \fover-40K. We use six benchmarks, including HANS~\citep{mccoy-etal-2019-right}, MMLU~\citep{mmlu-pro-nomath-2024}, and BBH~\citep{suzgun-etal-2023-challenging}. The results show that \fover-PRMs outperform baseline PRMs and achieve performance competitive with existing PRMs~\citep[e.g.,][]{xiong2024rlhflowmath, zhang2025lessons}, whose training data is constructed via human annotation or Monte Carlo roll-outs that require substantially higher cost than \fover.

Together, as shown in Figure~\ref{fig:performance-summary}, our experiments demonstrate the practical effectiveness of \fover{} in improving PRMs on widely used informal reasoning benchmarks. These results showcase formal-to-informal transfer and cross-task generalization in PRM training using formally verified labels annotated on formal reasoning tasks.

Our main contributions are as follows:
\begin{itemize}
    \item We propose \fover, a method that efficiently constructs accurate PRM training data from formal reasoning tasks using formal verification, addressing the annotation cost and labeling inaccuracy of existing approaches.
    \item Our experiments show the practical effectiveness of \fover, with PRM training data from formal reasoning tasks improving performance on informal reasoning tasks written in natural language, demonstrating formal-to-informal transfer and cross-task generalization in PRM training.
\end{itemize}

\section{Background and Related Work}

PRMs for reasoning tasks are typically created by fine-tuning LLMs on the classification task to detect errors in reasoning traces at the step level~\citep{lightman2024lets, wang-etal-2024-math-shepherd}. Training data for PRMs are often constructed by annotating step-level error labels on reasoning traces generated by LLMs. Given a problem $p \in \mathcal{P}$ from a reasoning task, an LLM generates a step-by-step solution consisting of $k$ steps:~$s = [s_1, s_2, \ldots, s_k] \sim \mathrm{LLM}(p)$. We then assign step-level error labels $[y_{s_1}, y_{s_2}, \ldots, y_{s_k}] \in \{0,1\}^k$, which serves as the supervision signal for PRM training. Here, we consider binary labels of correct and incorrect.

Despite extensive research on PRM training data construction, obtaining reliable step-level labels remains challenging. Existing approaches are costly and noisy, as they often rely on human annotation or computationally expensive automatic methods.

\paragraph{Human annotation.}
Human annotation~\citep{uesato2022solving, lightman2024lets} is the primary approach used in early studies. However, for step-level reasoning tasks, it is particularly costly, and achieving high inter-annotator agreement is difficult. For example, \citet{zheng2024processbench} discard 30\% of the annotated solutions due to low agreement.

\paragraph{Monte Carlo roll-outs.}
Monte Carlo roll-outs~\citep{wang-etal-2024-math-shepherd, luo2024improve}, also known as Math-Shepherd, is a widely used method for automating PRM training data creation. MC roll-outs generate multiple continuations from a target reasoning step and use the frequency of reaching the correct final answer as an estimate of the correctness of that step.

Specifically, to label a step $s_r$ in a solution $s = [s_1,\ldots,s_r,\ldots,s_k]$, the method samples multiple continuations from an LLM conditioned on the problem $p$ and the prefix $s_{1:r}$:
\begin{align*}
[s_1, \ldots, s_r, c^{(1)}_{r+1}, \ldots, c^{(1)}_{K_1}] &\sim \mathrm{LLM}\!\left(\,\cdot \mid p, s_{1:r}\right) \\
[s_1, \ldots, s_r, c^{(2)}_{r+1}, \ldots, c^{(2)}_{K_2}] &\sim \mathrm{LLM}\!\left(\,\cdot \mid p, s_{1:r}\right) \\
\ldots[s_1, \ldots, s_r, c^{(n)}_{r+1}, \ldots, c^{(n)}_{K_n}] &\sim \mathrm{LLM}\!\left(\,\cdot \mid p, s_{1:r}\right)
\end{align*}
Let $a_i$ denote the final answer produced by the $i$-th continuation and let $a^\ast$ be the ground-truth final answer. The quality of step $s_r$ is estimated as $q_{s_r} = \frac{1}{n} \sum_{i=1}^{n} \mathbb{I}\!\left(a_i = a^\ast\right)$, which can be used as a soft label or binarized to obtain a hard label.

Although widely adopted, MC roll-outs suffer from high computational cost due to the need to generate multiple continuations per step, and the estimated labels can be noisy and inaccurate.

\paragraph{Other approaches.}
Perturbations~\citep{yang-etal-2022-generating, ma2023letsrewardstep, paul-etal-2024-refiner} is another automated approach that introduces mistakes into generated solutions by applying heuristic methods or using LLMs. However, it results in training data with artificial and unnatural errors. Some work creates PRM training data by annotating error labels using stronger LLMs~\citep{zhang2025lessons, zeng2025versaprm}, but this approach results in merely distilling the capabilities of stronger LLMs.

\section{\fover{}}
\label{sec:fover-method}

We propose \fover, a method for creating PRM training data by annotating step-level error labels on formal reasoning tasks using formal verification tools. Using \fover, we create PRM training data named \fover-40K from first-order logical reasoning tasks using Z3 and from formal theorem proving tasks using Isabelle.

\subsection{Step-level Formal Verification}

\paragraph{Background: Formal verification.}
Formal reasoning tasks~(or symbolic reasoning tasks), including formal logic and formal theorem proving, are problems defined and solved using formal syntax and rules~\citep{nawaz2019survey, clark2021ruletakers}. 
Formal verification tools, such as automated solvers and theorem provers, are used to formally verify solutions to formal reasoning tasks~\citep{zhou2024dont, yang2024formal}. The tools differ in the methods they use and in the groups of tasks to which they apply. For example, Z3~\citep{demoura2008z3} is an SMT solver, which extends SAT solvers with background theories, applicable to decidable subsets of first-order logic. Isabelle/HOL~\citep{nipkow2002isabelle} is an interactive theorem prover applicable to higher-order logic.

To illustrate the process of formal verification, we present a simple example of a first-order logic reasoning~(logical entailment) task that can be verified using a SAT solver:
\begin{equation*}
\text{Context: } A \rightarrow B, \; B \rightarrow C, \; A 
\quad 
\text{Hypothesis: } C
\end{equation*}
Assume that the predicted answer is ``entailment,'' meaning the hypothesis is logically entailed by the context. We want to verify whether this prediction is correct, which is equivalent to showing that the following implication is valid:
\[
\big((A \rightarrow B) \wedge (B \rightarrow C) \wedge A\big) \;\rightarrow\; C .
\]
We can check this with a SAT solver using resolution: combine the premises with the negation of the hypothesis and test its satisfiability in Conjunctive Normal Form (CNF):
\[
(\lnot A \lor B) \wedge (\lnot B \lor C) \wedge A \wedge \lnot C.
\]
If the solver reports this formula unsatisfiable, then the entailment is confirmed. SAT solvers such as Z3 automate this process and thus can formally verify the correctness of the prediction.

\paragraph{Step-level formal verification.}
In this paper, we propose using formal verification tools to verify solutions at the step level. As illustrated above, these tools have primarily been applied to solution-level verification. Our key idea is that, for certain tasks, they can also be employed to verify the correctness of individual reasoning steps. For instance, for the above problem, suppose an LLM produces the following step-by-step solution:
\begin{align*}
\text{Step 1:}&\quad \big((A \rightarrow B) \wedge A\big) \;\rightarrow\; B.
\\ \text{Step 2:}&\quad \big(B \wedge (A \rightarrow B)) \;\rightarrow\; C.
\end{align*}

It can be verified at the step level: we adopt a simple strategy that evaluates the logical correctness of each step independently. Once we check that each step only uses the provided context and preceding results, we can verify the logical correctness by testing the satisfiability of:
\begin{align*}
\text{Step 1:}&\quad (\neg A \lor B)\wedge A \wedge \neg B.\\
\text{Step 2:}&\quad B \wedge (\neg A \lor B)\wedge \neg C.
\end{align*}
The SAT solver will show that the first formula is unsatisfiable, indicating Step~1 is correct, and the second is satisfiable, indicating Step~2 is incorrect.
This procedure provides step-level verification for the solution to the logical entailment task.

In this work, we implement step-level verification for formal logic~(logical entailment) using Z3 and theorem-proving tasks using Isabelle to create PRM training data.

\subsection{\fover{} Framework}
\begin{figure}[t]
    \centering
    \includegraphics[width=\linewidth]{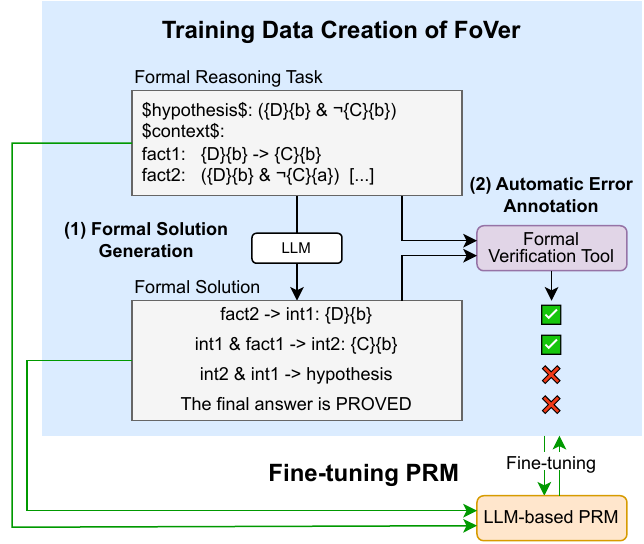}
    \caption{Top:~\fover{} creates PRM training data from formal reasoning tasks using formal verification tools. Bottom:~Using the training data, we fine-tune PRMs on the task of step-level verification.
    }
    \label{fig:fover-dataset-creation}
\end{figure}

\fover{} synthesizes PRM training data in two stages, as shown in Figure~\ref{fig:fover-dataset-creation}. In the first stage, given a problem $p \in \mathcal{P}$ from a formal reasoning task that is compatible with formal verification, an LLM generates a step-by-step formal solution consisting of $k$~steps:~$s = [s_1, s_2, \ldots, s_k] \sim \mathrm{LLM}(p)$. The solution $s$ may contain logical errors but is required to follow the format compatible with a formal verification tool. In this work, we provide a few-shot demonstration to guide LLMs. Alternatively, it is also possible to use models trained for specific formal verification tools~\citep{kaiyu2023leandojo, xin2024deepseekprover}. LLMs may generate solutions in an invalid format, so we generate multiple solutions until we obtain one with a valid format.

In the second stage, the formal verification tool assigns step-level error labels~$[y_{s_1}, y_{s_2}, \ldots, y_{s_k}] \in \{0, 1\}^K$, without the need for human annotation:
\begin{align*}
    &[ y_{s_1}, y_{s_2}, \ldots, y_{s_k} ] \\
    &\quad = \mathrm{FormalVerification}(p, [s_1, s_2, \ldots, s_k]),
\end{align*}
where $\mathrm{FormalVerification}(,)$, a formal verification tool, accurately outputs label $y_{s_k}=1$ if the solution step $s_k$ is correct, $y_{s_k}=0$ otherwise. 

Using training data created with \fover, we fine-tune an LLM on the step-level binary classification task to serve as a PRM~($\mathcal{P} \times \mathcal{S} \rightarrow [0,1]^K$), which is the standard approach for creating classification-based PRMs~\citep{zhang2025lessons}. The cross-entropy training objective for PRMs is given by:
\begin{equation*}
    \mathcal L = \sum_{i=1}^K \left( y_{s_i} \log r_{s_i} + (1 - y_{s_i}) \log (1 - r_{s_i})  \right),
\end{equation*}
where $r_{s_i}$ is the step-level score for step $s_i$ predicted by the PRM. In our implementation, we fine-tune LLM in a text generation task to output the token ``correct'' or ``incorrect'' for each step.

\begin{figure*}[t]
    \centering
    \includegraphics[width=\linewidth]{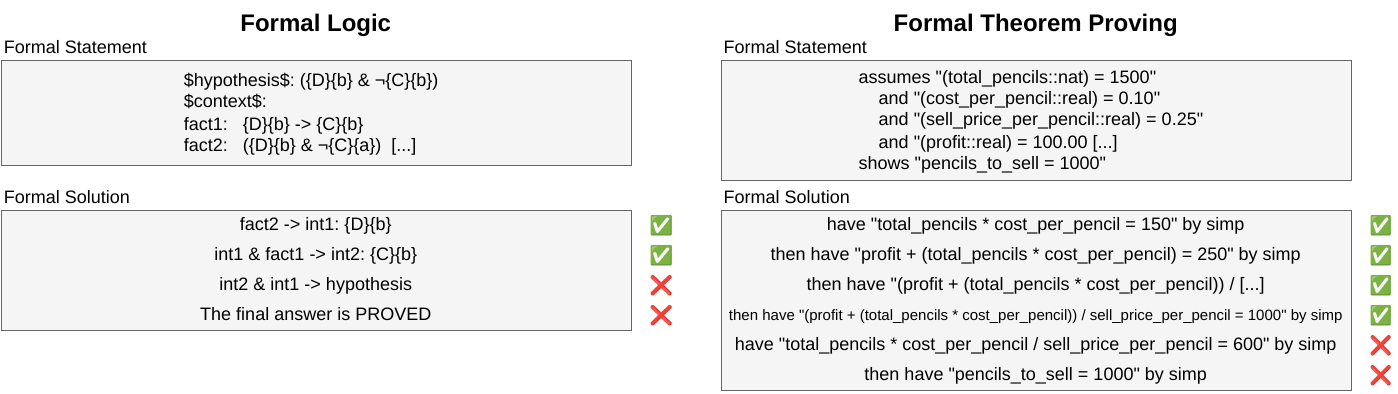}
    \caption{Examples from \fover-40K. 
    }
    \label{fig:fover-40k-examples}
\end{figure*}

\subsection{\fover-40K Dataset}
\label{sec:fover-40k}

Using \fover, we synthesize PRM training data that includes step-level binary error labels on the formal logic and formal theorem proving tasks, which we refer to as \fover-40K. As shown in Table~\ref{tab:fover40k-statistics}, \fover-40K includes 40K steps in reasoning traces generated by Llama~3.1~8B~\citep{dubey2024llama3} and Qwen~2.5~7B~\citep{qwen25} with error labels annotated by formal verification tools, Z3 and Isabelle. 
Figure~\ref{fig:fover-40k-examples} shows example data.

The \fover-40K dataset is created from FLDx2~\citep{morishita2023learning, morishita2024enhancing} and GSM8K-level mathematical reasoning tasks. We select these tasks based on diversity, simplicity, and generalization considerations. For the formal logic task, we selected FLDx2 because it offers the greatest diversity among those expressible via deduction rules~\citep{morishita2024enhancing}. For mathematical reasoning, we selected GSM8K-level datasets to simplify the verification pipeline.

\begin{table}[t]
    \setlength{\tabcolsep}{2.4pt}
    \centering
    \scriptsize
    \begin{tabular}{M{.12\linewidth}M{.19\linewidth}M{.15\linewidth}M{.18\linewidth}cc}
    \toprule
        \multirow{2.7}{*}{Tasks} &
        \multirow{2.7}{\linewidth}{\centering Source Datasets} &
        \multirow{2.7}{*}{\parbox{\linewidth}{\centering Formal Verification Tools}} &
        \multirow{2.7}{*}{\parbox{\linewidth}{\centering Step-by-step Solutions}} &
        \multicolumn{2}{c}{Error Labels} \\
        \cmidrule(l{2pt}r{2pt}){5-6}
        & & & & \# of Steps & \% Error \\
    \midrule
        \multirow{2}{\linewidth}{\centering Formal Logic}
        & \multirow{2}{*}{FLDx2} & \multirow{2}{*}{Z3} & Llama 3.1 8B & 10,000 & 50\% \\
        &  &  & Qwen 2.5 7B & 10,000 & 50\% \\
    \midrule
        \multirow{3}{\linewidth}{\centering Formal Theorem Proving}
        & \multirow{3}{*}{\parbox{\linewidth}{\centering GSM8K, MetaMathQA, Big-Math}} & \multirow{3}{*}{Isabelle} & \multirow{3}{*}{Qwen 2.5 7B} & \multirow{3}{*}{20,000} & \multirow{3}{*}{50\%} \\
        &  &  &  &  &  \\
        &  &  &  &  &  \\
    \bottomrule
    \end{tabular}
    \caption{Statistics of the \fover-40K training dataset. 
    }
    \label{tab:fover40k-statistics}
\end{table}

\paragraph{Formal logic.}
We use the logical entailment task in FLDx2~\citep{morishita2023learning, morishita2024enhancing}, a dataset for multi-step first-order logic deduction, in which the goal is to determine whether a hypothesis is entailed by a given set of premises. We generate step-by-step formal solutions with the LLMs and annotate step-level error labels using Z3.

\paragraph{Formal theorem proving.}
We use the task of formal theorem proving for verifying solutions to math word problems~\citep{wu2022autoformalization, zhou2024dont}. The conditions of a math word problem are presented as premises, a candidate final answer is formulated as a hypothesis, and the task is to generate a proof for this statement. We generate formal proofs with the LLMs in the task of verifying the solutions to GSM8K-level problems, including GSM8K~\citep{cobbe2021gsm8k}, GSM8K-based cases in MetaMathQA~\citep{yu2024metamath}, and math word problems in Big-Math~\citep{albalak2025bigmath}. We then annotate step-level error labels using Isabelle.

\begin{table*}[t]
\centering
\scriptsize
\begin{subtable}{\linewidth}
\centering
\begin{tabular}{M{.1\linewidth}ccccc}
\toprule
& \multicolumn{2}{c}{\multirow{2.7}{*}{PRMs}} & \multicolumn{2}{c}{Logic} & \multirow{2.7}{*}{Average} \\
\cmidrule(l{2pt}r{2pt}){4-5}
& & & FOLIO & LogicNLI &  \\
\midrule
\multirow{2}{\linewidth}{\centering PRMs~based~on Llama~3.1~8B}
& \multicolumn{2}{l}{Llama 3.1 8B} & 58.6\phantom{$^*$} & 39.2\phantom{$^*$} & 48.9\phantom{$^*$} \\
& \multicolumn{2}{l}{\fover-Llama3.1-8B} & {\bf 60.1}\phantom{$^*$} & {\bf 41.2}\phantom{$^*$} & {\bf 50.6}\phantom{$^*$} \\
\midrule
\multirow{2}{\linewidth}{\centering PRMs~based~on Qwen~2.5~7B}
& \multicolumn{2}{l}{Qwen 2.5 7B} & {\bf 64.0}\phantom{$^*$} & 44.4\phantom{$^*$} & 54.2\phantom{$^*$} \\
& \multicolumn{2}{l}{\fover-Qwen2.5-7B} & 63.5\phantom{$^*$} & {\bf 49.6}\phantom{$^*$} & {\bf 56.6}\phantom{$^*$} \\
\bottomrule
\end{tabular}
\hskip 1em
\begin{tabular}{ccccc}
\toprule
\multicolumn{4}{c}{Math} & \multirow{2.7}{*}{Average} \\
\cmidrule(l{2pt}r{2pt}){1-4}
GSM8K & MATH & AQuA & AIME & \\
\midrule
87.6\phantom{$^*$} & {\bf 54.4}\phantom{$^*$} & 66.1\phantom{$^*$} & 4.0\phantom{$^*$}  & 53.0\phantom{$^*$} \\
{\bf 88.4}\phantom{$^*$} & 53.6\phantom{$^*$} & {\bf 71.7}$^*$           & {\bf 5.2}\phantom{$^*$}  & {\bf 54.7}$^*$           \\
\midrule
90.0\phantom{$^*$} & 76.4\phantom{$^*$} & 80.3\phantom{$^*$} & {\bf 12.4}\phantom{$^*$} & 64.8\phantom{$^*$} \\
{\bf 92.4}\phantom{$^*$} & {\bf 79.2}\phantom{$^*$} & {\bf 81.1}\phantom{$^*$} & 12.0\phantom{$^*$} & {\bf 66.2}$^*$           \\
\bottomrule
\end{tabular}
\end{subtable}
\caption{Best-of-K~(K=7) performance on informal logic and math benchmarks. The first rows represent the baseline PRMs based on LLMs without additional training. \fover-PRMs exhibit significantly better performance, showing that \fover-40K improves verification on informal logic and math reasoning. $^*$:~Statistically significant improvement over the baseline PRMs in the first row~($p<0.05$, paired bootstrap).}
\label{tab:best-of-k-math-logic}
\end{table*}

\section{Experimental Results} \label{sec:experiments}

This section evaluates our \fover-PRMs fine-tuned on \fover-40K. While \fover-40K includes step-level error labels on formal reasoning tasks, PRMs are often used to detect mistakes in informal reasoning tasks written in natural language. Therefore, we evaluate \fover-PRMs on widely used informal reasoning benchmarks to answer the following research questions.

\begin{itemize}
    \item RQ1: Does training on \fover-40K improve PRMs on informal logic and math reasoning tasks, which are informal variants of the training tasks?~(formal-to-informal transfer, \S\ref{sec:results-math-logic})
    \item RQ2: Does training on \fover-40K improve PRMs on informal reasoning tasks that largely differ from the training tasks, such as NLI and BBH?~(cross-task generalization, \S\ref{sec:experiments-unseen})
\end{itemize}

\subsection{Experimental Setting}

\paragraph{Evaluation methodology.}
We evaluate PRMs using the Best-of-K performance on reasoning benchmarks and step-level verification performance, both of which are widely used approaches to assess PRMs~\citep{li-etal-2023-making, zhang2024generative}.
First, \textbf{Best-of-K}~\citep{cobbe2021gsm8k, li-etal-2023-making} uses PRMs to select the best solution from multiple candidates generated by LLMs for the same input. The performance of the selected solutions indirectly indicates the verification performance of the PRMs. We use few-shot demonstrations to generate $K=7$ step-by-step solutions from Llama~3.1~8B and Qwen~2.5~7B with a temperature of 0.5. Following prior work~\citep{lightman2024lets, wang-etal-2024-math-shepherd}, we then use PRMs to score each step, and the solution score is the minimum score across all steps. The solution with the highest solution score is selected. For large datasets, we use 250 randomly sampled examples from each dataset. Second, we evaluate \textbf{step-level verification} performance on math reasoning tasks in ProcessBench~\citep{zheng2024processbench}, which includes human annotated binary step-level error labels. ProcessBench includes labels only for the earliest error in each reasoning traces, so we use the steps up to the first error in each solution.

\paragraph{Evaluated PRMs.}
We evaluate classification-based PRMs that use Llama~3.1~8B~\citep{dubey2024llama3} and Qwen~2.5~7B~\citep{qwen25} as the backbone. We choose these LLMs because they are widely used as backbones for PRMs, which allows for direct comparison with PRMs introduced in prior work. Our PRMs, referred to as {\bf \fover-PRMs}, are obtained by fine-tuning these LLMs on \fover-40K.
{\bf Baseline PRMs:}~We compare with baselines that use the backbone LLMs as PRMs without additional fine-tuning. {\bf Existing PRMs:}~In Section~\ref{sec:experiments-unseen}, we also evaluate five PRMs introduced in prior work~(Table~\ref{tab:evaluted-prms}). Among PRMs based on Llama~3.1~8B, we evaluate RLHFlow-Llama3.1-8B trained on the DeepSeek or Mistral data~\citep{xiong2024rlhflowmath}, which include error labels on solutions generated by stronger models on GSM8K and MATH acquired via Monte Carlo roll-outs~\citep{wang-etal-2024-math-shepherd}. Among PRMs based on Qwen~2.5~7B, we evaluate Qwen2.5-Math-7B-PRM800K~\citep{zheng2024processbench}, which is trained on human-annotated labels on MATH, and Qwen2.5-Math-PRM-7B~\citep{zhang2025lessons}, which is trained on labels synthesized using Monte Carlo roll-outs and verification by a stronger model. We also evaluate Qwen2.5-7B-Skywork-PRM~\citep{skyworkopeno12024}, which is trained on math and coding.

\newcommand{\directevaluationheaderfirst}{
\multicolumn{4}{c}{ProcessBench} & \multirow{2.7}{*}{Average} \\
}
\newcommand{\directevaluationheadermidrule}[1]{
#1 \cmidrule(l{2pt}r{2pt}){3-6}
}
\newcommand{\directevaluationheadersecond}{
GSM8K & MATH & Olympiad & Omni
}

\begin{table}[t]
\setlength{\tabcolsep}{3pt}
\centering
\scriptsize
\begin{subtable}{\linewidth}
\begin{tabular}{cccccccc}
\toprule
\multicolumn{2}{c}{PRMs} & \directevaluationheadersecond & Ave. \\
\midrule
\multicolumn{2}{l}{Llama 3.1 8B} & 70.0\phantom{$^*$} & 67.4\phantom{$^*$} & 68.3\phantom{$^*$} & 63.9\phantom{$^*$} & 67.4\phantom{$^*$} \\
\multicolumn{2}{l}{\fover-Llama3.1-8B-PRM} & {\bf 81.2}$^*$           & {\bf 76.4}$^*$           & {\bf 75.5}$^*$           & {\bf 76.3}$^*$           & {\bf 77.3}$^*$           \\
\midrule
\multicolumn{2}{l}{Qwen 2.5 7B} & 75.5\phantom{$^*$} & 78.2\phantom{$^*$} & 76.2\phantom{$^*$} & 73.6\phantom{$^*$} & 75.9\phantom{$^*$} \\
\multicolumn{2}{l}{\fover-Qwen2.5-7B-PRM} & {\bf 86.6}$^*$           & {\bf 88.0}$^*$           & {\bf 84.0}$^*$           & {\bf 84.5}$^*$           & {\bf 85.8}$^*$           \\
\bottomrule
\end{tabular}
\end{subtable}
\caption{Step-level binary classification performance of PRMs on math reasoning tasks in ProcessBench~(AUROC). \fover-PRMs exhibit significantly better performance than the baselines, showing that \fover-40K improves verification on informal math reasoning. $^*$:~Statistically significant improvement over the baseline PRMs in the first rows~($p<0.05$, paired bootstrap).}
\label{tab:processbench-result}
\end{table}

\begin{table*}[t]
    \setlength{\tabcolsep}{3.5pt}
    \centering
    \scriptsize
    \begin{tabular}{cM{.16\linewidth}M{.14\linewidth}M{.13\linewidth}M{.12\linewidth}M{.07\linewidth}M{.07\linewidth}}
    \toprule
        PRM                        & Tasks in Training Data               & Step-Level Error Annotation & \# LLM Calls per Solution for Labeling  & Does Not Require Stronger LLMs & No Human Annotation & Accurate Annotation \\
    \midrule
        RLHFlow-Llama3.1-8B (M, D) & Math Reasoning                       & MC Roll-out           & 8 $\times$ \# Steps & \cmark & \cmark & \xmark \\
    \midrule
        Qwen2.5-Math-7B-PRM800K    & Math Reasoning                       & Human Annotation      & 1                   & \cmark & \xmark & \xmark \\
        Qwen2.5-7B-Skywork-PRM     & Math Reasoning, Coding               & MC Roll-out           & (not reported)      & \cmark & \cmark & \xmark \\
        Qwen2.5-Math-PRM-7B        & Math Reasoning                       & MC Roll-out, LLMs     & 8 $\times$ \# Steps & \xmark & \cmark & \xmark \\
    \midrule
        \fover-PRM~(Ours)           & Formal~Logic, Formal~Theorem~Proving & Formal Verification   & 1                   & \cmark & \cmark & \cmark \\
    \bottomrule
    \end{tabular}
    \caption{Training data creation methods for PRMs evaluated in Section~\ref{sec:experiments-unseen}.
    }
    \label{tab:evaluted-prms}
\end{table*}

\begin{table*}[t!]
\setlength{\tabcolsep}{5pt}
\scriptsize
\begin{subtable}{\linewidth}
\centering
\begin{tabular}{M{.1\linewidth}ccccc}
\toprule
& \multicolumn{2}{c}{\multirow{2.7}{*}{PRMs}} & \multicolumn{2}{c}{NLI} & \multirow{2.7}{*}{Average} \\
\cmidrule(l{2pt}r{2pt}){4-5}
& & & ANLI & HANS &  \\
\midrule
\multirow{5.4}{\linewidth}{\centering PRMs~based~on Llama~3.1~8B}
& \multicolumn{2}{l}{Llama 3.1 8B} & 26.4\phantom{$^*$} & 69.2\phantom{$^*$} & 47.8\phantom{$^*$} \\
\cmidrule{2-6}
& \multicolumn{2}{l}{RLHFlow-Llama3.1-8B-M} & 26.8\phantom{$^*$} & 77.6$^*$           & 52.2$^*$           \\
& \multicolumn{2}{l}{RLHFlow-Llama3.1-8B-D} & 27.2\phantom{$^*$} & 74.4$^*$           & 50.8$^*$           \\
\cmidrule{2-6}
& \multicolumn{2}{l}{\fover-Llama3.1-8B-PRM (ours)} & {\bf 30.0}\phantom{$^*$} & {\bf 81.2}$^*$           & {\bf 55.6}$^*$           \\
\midrule
\midrule
\multirow{6.4}{\linewidth}{\centering PRMs~based~on Qwen~2.5~7B}
& \multicolumn{2}{l}{Qwen 2.5 7B} & {\bf 32.4}\phantom{$^*$} & 84.4\phantom{$^*$} & {\bf 58.4}\phantom{$^*$} \\
\cmidrule{2-6}
& \multicolumn{2}{l}{Qwen2.5-Math-7B-PRM800K} & 29.6\phantom{$^*$} & 83.2\phantom{$^*$} & 56.4\phantom{$^*$} \\
& \multicolumn{2}{l}{Qwen2.5-7B-Skywork-PRM} & 29.2\phantom{$^*$} & 81.2\phantom{$^*$} & 55.2\phantom{$^*$} \\
& \multicolumn{2}{l}{Qwen2.5-Math-PRM-7B} & 27.6\phantom{$^*$} & {\bf 86.0}\phantom{$^*$} & 56.8\phantom{$^*$} \\
\cmidrule{2-6}
& \multicolumn{2}{l}{\fover-Qwen2.5-7B-PRM (ours)} & 30.8\phantom{$^*$} & {\bf 86.0}\phantom{$^*$} & {\bf 58.4}\phantom{$^*$} \\
\bottomrule
\end{tabular}
\hskip 1em
\begin{tabular}{c}
\toprule
MMLU \\
\cmidrule(l{2pt}r{2pt}){1-1}
Pro-NoMath \\
\midrule
56.8\phantom{$^*$} \\
\midrule
56.8\phantom{$^*$} \\
{\bf 57.6}\phantom{$^*$} \\
\midrule
54.8\phantom{$^*$} \\
\midrule
\midrule
58.0\phantom{$^*$} \\
\midrule
{\bf 62.0}$^*$           \\
58.8\phantom{$^*$} \\
60.4\phantom{$^*$} \\
\midrule
60.4\phantom{$^*$} \\
\bottomrule
\end{tabular}
\hskip 1em
\begin{tabular}{cccc}
\toprule
\multicolumn{3}{c}{BBH} & \multirow{2.7}{*}{Average} \\
\cmidrule(l{2pt}r{2pt}){1-3}
Temporal & Tracking & Sorting &  \\
\midrule
90.4\phantom{$^*$} & 90.0\phantom{$^*$} & 41.6\phantom{$^*$} & 74.0\phantom{$^*$} \\
\midrule
92.0\phantom{$^*$} & 90.8\phantom{$^*$} & 39.2\phantom{$^*$} & 74.0\phantom{$^*$} \\
{\bf 98.8}$^*$           & 92.8\phantom{$^*$} & 40.0\phantom{$^*$} & 77.2$^*$           \\
\midrule
97.2$^*$           & {\bf 96.0}$^*$           & {\bf 46.8}$^*$           & {\bf 80.0}$^*$           \\
\midrule
\midrule
91.2\phantom{$^*$} & 89.2\phantom{$^*$} & 28.8\phantom{$^*$} & 69.7\phantom{$^*$} \\
\midrule
82.0\phantom{$^*$} & 90.8\phantom{$^*$} & 28.4\phantom{$^*$} & 67.1\phantom{$^*$} \\
84.0\phantom{$^*$} & 91.2\phantom{$^*$} & {\bf 31.6}\phantom{$^*$} & 68.9\phantom{$^*$} \\
86.4\phantom{$^*$} & {\bf 92.8}$^*$     & 28.8\phantom{$^*$} & 69.3\phantom{$^*$} \\
\midrule
{\bf 92.4}\phantom{$^*$} & 90.0\phantom{$^*$} & 29.6\phantom{$^*$} & {\bf 70.7}\phantom{$^*$} \\
\bottomrule
\end{tabular}
\end{subtable}
\caption{Best-of-K~(K=7) performance on tasks that differ from tasks in training data of \fover-PRMs and existing PRMs~(i.e., unseen tasks). The first rows represent the baseline PRMs based on LLMs without additional training. The next set of rows contains existing PRMs introduced in prior work. The last rows show our \fover-PRMs. $^*$:~Statistically significant improvement over the baseline PRMs in the first row ($p<0.05$, paired bootstrap).}
\label{tab:best-of-k-unseen}
\end{table*}

\subsection{Results on Logic and Math}
\label{sec:results-math-logic}

First, we evaluate \fover-PRMs on widely-used informal logic and math reasoning benchmarks, which are informal variants of formal logic and theorem proving tasks included in \fover-40K.

\paragraph{Results of Best-of-K.}
We evaluate the Best-of-K performance of PRMs on two logical reasoning benchmarks:~FOLIO~\citep{han-etal-2024-folio}, LogicNLI~\citep{tian-etal-2021-diagnosing}, and four math benchmarks:~GSM8K~\citep{cobbe2021gsm8k}, MATH~\citep{hendrycks2021measuring}, AQuA-RAT~\citep{ling-etal-2017-program}, AIME~(2016-2024)~\citep{aops_aime}.
Table~\ref{tab:best-of-k-math-logic} reports the Best-of-K performance, showing that \fover-PRMs outperform the baseline PRMs on average both on the informal logic and math tasks.

\paragraph{Results on ProcessBench.}
We also evaluate the PRMs on the step-level binary classification~(error detection) task on mathematical reasoning tasks in ProcessBench~\citep{zheng2024processbench}. ProcessBench includes human-annotated step-level error labels (correct vs.\ incorrect) for reasoning traces on four math benchmarks:~GSM8K, MATH, Olympiad-Bench, and Omni-Math.
Table~\ref{tab:processbench-result} shows the step-level verification performance of PRMs, measured in AUROC. The results show that fine-tuning on \fover-40K significantly improves PRM performance at detecting step-level mistakes from the baseline PRMs in informal math reasoning tasks.

\begin{takeawaybox}{{\fontsize{10}{10}\selectfont Answer to RQ1: Formal-to-Informal Transfer}}
PRMs trained on \fover-40K outperform baseline PRMs on average on informal math and logic benchmarks. These results demonstrate that PRM training on formally verified error labels on formal reasoning tasks exhibits formal-to-informal transfer and improves performance on informal variants of the formal training tasks.
\end{takeawaybox}

\newcommand{\colorcircle}[1]{
    \begin{tikzpicture}
        \filldraw[draw=none,fill=#1] (0,0) circle (.7ex);
    \end{tikzpicture}
}
\newcommand{\colortriangle}[1]{
    \begin{tikzpicture}
        \filldraw[draw=none,fill=#1] (0,0) -- (1.2ex,0) -- (0.6ex,1.2ex) -- cycle;
    \end{tikzpicture}
}
\newcommand{\invertedcolortriangle}[1]{
    \begin{tikzpicture}
        \filldraw[draw=none,fill=#1] (0,1.2ex) -- (1.2ex,1.2ex) -- (0.6ex,0) -- cycle;
    \end{tikzpicture}
}

\newcommand{\baselinecircle}{\colorcircle{skyblue}}

\newcommand{\wronggt}{%
  \begin{tikzpicture}[baseline=-.5ex, line cap=round]
    \fill[Gray] (0,0) circle (.7ex);
    \begin{scope}
      \clip (0,0) circle (.7ex);
      \draw[white, line width=1pt] (-0.8ex,-0.6ex) -- ++(1.0ex,1.2ex);
      \draw[white, line width=1pt] (-0.4ex,-1.0ex) -- ++(1.2ex,1.4ex);
    \end{scope}
  \end{tikzpicture}%
}

\newcommand{\wrongintermediate}{%
  \begin{tikzpicture}[baseline=-.5ex]
    \fill[Gray] (0,0) circle (.7ex);
    \begin{scope}
      \clip (0,0) circle (.7ex);
      \foreach \x in {-0.28ex,0.28ex}
        \foreach \y in {-0.28ex,0.28ex}
          \fill[white] (\x,\y) circle (0.18ex);
    \end{scope}
  \end{tikzpicture}%
}

\newcommand{\bothwrong}{%
  \begin{tikzpicture}[baseline=-.5ex, line width=1pt]
    \fill[Gray] (0,0) circle (.7ex);
    \begin{scope}
      \clip (0,0) circle (.7ex);
      \draw[white] (-0.5ex,-0.5ex) -- (0.5ex,0.5ex);
      \draw[white] (-0.5ex,0.5ex)  -- (0.5ex,-0.5ex);
    \end{scope}
  \end{tikzpicture}%
}

\subsection{Results on Unseen Tasks} \label{sec:experiments-unseen}

Second, we evaluate \fover-PRMs on reasoning tasks that are substantially different from the formal logic and theorem proving tasks in \fover-40K, which we refer to as unseen tasks. We conduct evaluations on 6~reasoning benchmarks spanning three tasks: NLI, including ANLI~\citep{nie-etal-2020-adversarial} and HANS~\citep{mccoy-etal-2019-right}; MMLU-Pro-NoMath~\citep{mmlu-pro-nomath-2024}; and BIG-Bench Hard, including temporal sequences, tracking shuffled objects (three objects), and word sorting~\citep{suzgun-etal-2023-challenging}.
Here, in addition to \fover-PRMs, we evaluate existing PRMs introduced in prior work~(Table~\ref{tab:evaluted-prms}). Since these PRMs are trained on mathematical reasoning or coding tasks, the reasoning benchmarks evaluated in this section are also unseen for the existing PRMs.

Table~\ref{tab:best-of-k-unseen} shows the Best-of-K performance of PRMs on the unseen tasks. First, compared to the baseline PRMs, \fover-PRMs improve performance on average on almost all tasks. Furthermore, \fover-PRMs are competitive with existing PRMs, whose training datasets are created at substantially higher cost. In particular, \fover-Llama3.1-8B-PRM achieves the best performance on the majority of the benchmarks, compared with existing PRMs based on the same LLM.

\begin{takeawaybox}{{\fontsize{10}{10}\selectfont Answer to RQ2: Cross-Task Generalization}}
PRMs trained on formal reasoning tasks in \fover-40K exhibit cross-task generalization, improving performance on informal reasoning tasks such as NLI and BBH that are substantially different from the formal logic and theorem proving tasks in \fover-40K.
\end{takeawaybox}

\section{Analysis} \label{sec:analysis}

\paragraph{Ablation on training tasks.}
\fover-40K consists of formal logic and theorem proving tasks. To assess the contribution of each task, we train PRMs on each task separately and evaluate their performance. Table~\ref{tab:ablation-tasks} reports the Best-of-K performance of PRMs based on Llama~3.1~8B trained on different subsets of \fover-40K. The results show that each task is independently effective and leads to significant improvements over the baseline PRMs on average. Looking more closely, we observe that combining the two tasks during training improves performance on unseen tasks~(HANS, Temporal, and Tracking), while PRMs trained on individual tasks achieve better performance on their corresponding logic and math benchmarks. In light of this observation, when creating PRMs intended for use in a specific domain with available in-distribution training data, it may be preferable not to include additional domains in the training data. We will add this discussion to the updated version of the paper.

\begin{table}[t]
\centering
\fontsize{7}{7}\selectfont
\setlength{\tabcolsep}{2pt}
\begin{tabular}{M{.1\linewidth}M{.12\linewidth}cccccccc}
\toprule
Formal Logic & Formal Theorem & GSM8K & LogicNLI & HANS & Temporal & Tracking & Ave. \\
\midrule
--     & --     & 87.6\phantom{$^*$} & 39.2\phantom{$^*$} & 69.2\phantom{$^*$} & 90.4\phantom{$^*$} & 90.0\phantom{$^*$} & 75.3\phantom{$^*$} \\
\midrule
\cmark & --     & 86.8\phantom{$^*$} & {\bf 46.0}$^*$           & 75.2$^*$           & 96.8$^*$           & 92.4\phantom{$^*$} & 79.4$^*$           \\
--     & \cmark & {\bf 90.4}\phantom{$^*$} & 44.0\phantom{$^*$} & 76.8$^*$           & 96.0$^*$           & 94.8$^*$           & 80.4$^*$           \\
\midrule
\cmark & \cmark & 88.4\phantom{$^*$} & 41.2\phantom{$^*$} & {\bf 81.2}$^*$           & {\bf 97.2}$^*$           & {\bf 96.0}$^*$           & {\bf 80.8}$^*$           \\
\bottomrule
\end{tabular}
\caption{
Ablation on training tasks for Best-of-K~(K=7) performance of PRMs based on Llama~3.1~8B trained on different subsets of \fover-40K.
The top row represents the baseline PRM, and the last row represents the PRM trained on the original \fover-40K.
}
\label{tab:ablation-tasks}
\end{table}

\begin{table}[t]
\centering
\fontsize{7}{7}\selectfont
\setlength{\tabcolsep}{1.8pt}
\begin{tabular}{ccccccccc}
\toprule
PRMs & GSM8K & LogicNLI & HANS & Temporal & Tracking & Ave. \\
\midrule
Qwen~3~4B        & {\bf 93.2}\phantom{$^*$} & 66.4\phantom{$^*$} & {\bf 89.6}\phantom{$^*$} & 87.6\phantom{$^*$} & {\bf 95.6}\phantom{$^*$} & 86.5\phantom{$^*$} \\
\fover-Qwen3-4B & 91.6\phantom{$^*$} & {\bf 70.0}\phantom{$^*$} & {\bf 89.6}\phantom{$^*$} & {\bf 89.2}\phantom{$^*$} & {\bf 95.6}\phantom{$^*$} & {\bf 87.2}\phantom{$^*$} \\
\bottomrule
\end{tabular}
\caption{Best-of-K~(K=7) performance of PRMs based on Qwen~3~4B. Fine-tuning on \fover-40K~(second row) yields improved or comparable performance on all but one benchmark.
}
\label{tab:analysis-backbone}
\end{table}

\paragraph{Robustness to stronger backbone LLM.}
To evaluate the robustness of PRM training with \fover, we additionally evaluate PRMs based on Qwen~3~4B~\citep{qwen3}, which is newer and stronger than Llama~3.1~8B and Qwen~2.5~7B. Table~\ref{tab:analysis-backbone} reports the Best-of-K performance of PRMs based on Qwen~3~4B with and without fine-tuning on \fover-40K. The results show that fine-tuning on \fover-40K yields improved or comparable performance across nearly all benchmarks, aside from GSM8K. These findings indicate that \fover-40K, which is constructed from responses generated by weaker models, improves PRMs built on stronger LLM backbones, demonstrating the robustness of training with \fover{} with respect to backbone LLM choice.

\begin{figure}[t]
    \centering
    \includegraphics[width=\linewidth]{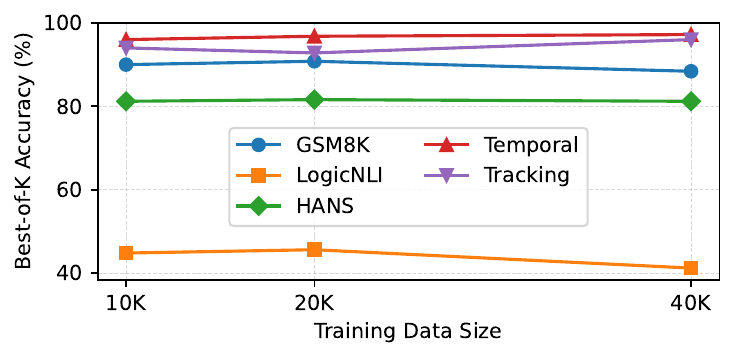}
    \caption{Ablation on training data size for PRMs based on Llama~3.1~8B trained on randomly sampled subsets of \fover-40K with 10K and 20K instances.}
    \label{fig:ablation-size}
\end{figure}

\paragraph{Ablation on training data size.}
Figure~\ref{fig:ablation-size} presents an ablation study on training data size for the Best-of-K performance of PRMs based on Llama~3.1~8B trained on randomly sampled subsets of \fover-40K with 10K and 20K instances. The results show that PRMs trained on a subset with 10K instances already achieve performance comparable to those trained on the full dataset, showing the data efficiency in the \fover{} training.

\begin{table}[t!]
\fontsize{8}{8}\selectfont
\setlength{\tabcolsep}{3pt}
\begin{subtable}{\linewidth}
\centering
\begin{mdframed}
{\bf Problem:} Today, Sarah went to the art show. Between what times could they have gone? [...] Lisa saw Sarah working at the office from 11am to 2pm. [...] Options: (A) 11am to 2pm (B) 7pm to 9pm [...]

{\bf Ground truth answer:} (B) 7pm to 9pm
\end{mdframed}
\vskip .5em
\begin{tabular}{cL{.58\linewidth}cc}
\toprule
Step & \multicolumn{1}{c}{Solution from Llama 3.1 8B} & Baseline & \fover \\
\midrule
2 & Lisa saw Sarah working at the office from 11am to 2pm, so Sarah could have gone to the art show from 2pm. &
0.98 & 0.99 \\
\midrule
7 & \textcolor{red}{From the options, we can see that 11am to 2pm is the only option that fits this time frame.} & 
{\bf 0.94} & {\bf 0.06} \\
\bottomrule
\end{tabular}
\caption{Positive example of \fover-PRM on the temporal sequence task. \fover-Llama3.1-8B-PRM detects an error in Step~7, whereas the baseline erroneously assigns a high score.} \label{tab:positive-example}
\end{subtable}
\vskip 1.5em
\begin{subtable}{\linewidth}
\centering
\begin{mdframed}
{\bf Problem:} For what integer $x$ is $\frac 35 < \frac x7 < \frac 79$?
\end{mdframed}
\vskip .5em
\begin{tabular}{cL{.58\linewidth}cc}
\toprule
Step & \multicolumn{1}{c}{Solution from Llama 3.1 8B} & Baseline & \fover \\
\midrule
0 & We can rewrite the inequality as $\frac{3}{5} < \frac{x}{7} < \frac{7}{9}$. &
1.00 & 1.00 \\
\midrule
1 & Multiplying the inequality by 315 to eliminate the fractions, we obtain $189 < 45x < 245$. &
1.00 & 0.98 \\
\midrule
2 & Dividing the inequality by 45, we get $\frac{189}{45} < x <$ \textcolor{red}{$\frac{245}{45}$} $\Rightarrow 4.2 < x < $ \textcolor{red}{$5.44$}. &
{\bf 1.00} & {\bf 0.68} \\
\bottomrule
\end{tabular}
\caption{Negative example for \fover-PRM on MATH. \fover-Llama3.1-8B-PRM is too strict in Step~2 and considers rounding $5.44444\ldots$ to $5.44$ as a mistake.} \label{tab:negative-example}
\end{subtable}
\caption{Example outputs from \fover-PRMs.}
\label{tab:case-study}
\end{table}

\paragraph{Case study.}
Table~\ref{tab:case-study} shows example outputs from \fover-PRM and baseline PRMs. Table~\ref{tab:positive-example} shows a {\bf positive example} on a solution to the temporal sequence task in BBH. The task is to find a free time slot that does not overlap with any events in a given list of past time intervals, which is distinct from formal reasoning tasks in \fover-40K. This example shows that \fover-Llama3.1-8B-PRM properly understands the task and solutions described in natural language and correctly detects a logical error in Step~7, showing the formal-to-informal transfer and cross-task generalization in PRM training with \fover. Table~\ref{tab:negative-example} shows a {\bf negative example} illustrating a potential side effect of \fover. In this example, \fover-Llama3.1-8B-PRM assigns a low score when a solution step rounds $\frac{245}{45} = 5.44444\ldots$ to $5.44$ for conciseness. While this can be considered a mistake under strict evaluation, it is often acceptable when applied to intermediate values where the exact precision is not required for the correctness of the final result. Training on formally annotated labels in \fover{} may cause PRMs to become overly strict, assigning lower scores to intermediate steps that are acceptable but not strictly precise.

\section{Discussion}

Our experiments show that PRMs trained on symbolic solutions for formal tasks improve performance on widely used informal reasoning tasks, demonstrating effective formal-to-informal transfer in PRM training. Compared to prior work that directly fine-tunes LLMs on symbolic reasoning traces, our results indicate that PRM training yields stronger transferability. For instance, \citet{morishita2024enhancing} train LLMs on reasoning traces from a formal logic dataset to improve general reasoning, but report limited transfer in a naive setup. To address this issue, they convert symbolic solutions into natural language and apply RecAdam~\citep{chen-etal-2020-recall}. In contrast, our results show that PRM training exhibits formal-to-informal transfer without these modifications. We attribute this advantage to the consistency of the output space in PRM training. Regardless of the input format, PRMs generate a binary label indicating whether each step is correct or incorrect, which reduces shifts in model behavior that can arise when LLMs are directly fine-tuned to generate symbolic reasoning traces.

\section{Conclusion}

We introduce \fover, a method for constructing PRM training data from formal reasoning tasks by annotating step-level error labels using formal verification tools. In contrast to existing approaches that are costly and noisy, \fover{} efficiently produces accurate PRM training data without relying on human annotation or repeated LLM calls. Experimental results show that PRMs trained on data generated by \fover{} achieve improved performance across reasoning benchmarks in math, logic, NLI, and BBH, demonstrating informal-to-formal transfer and cross-task generalization in PRM training. These findings show that formal verification provides a practical approach to PRM training data construction, improving PRMs on widely used informal reasoning tasks written in natural language.

\section*{Limitations} \label{appendix:limitations}

This work focuses on classification-based PRMs that generate a single score to each step. Although recent studies introduce PRMs that provide step-level reasoning for error classification~\citep{khalifa2025process, feng2025prmnecessary, kim2025scaling}, we leave the study of this type of PRM to future work, as they are not directly comparable with classification-based PRMs. PRMs with step-level reasoning incur substantially higher computational costs for verification, often multiple times greater than those required to generate the target reasoning traces, which limits their practical usefulness.

\ifreview\else
\section*{Acknowledgments}

This work was supported by NSF IIS-2338418 and DMS-2533995. We thank Jin Peng Zhou for providing guidance on the use of his code~\citep{zhou2024dont}. We also thank Terufumi Morishita for the valuable discussions and for his assistance with the FLDx2 dataset~\citep{morishita2024enhancing}, and we appreciate NLP Colloquium JP for providing the opportunity to connect with him. We thank Hieu N.\ Nguyen for his suggestions on improving the presentation of this work.

\fi

\bibliography{custom}

\appendix

\newpage
\let\addcontentsline\oldaddcontentsline

\setcounter{tocdepth}{2}
\renewcommand{\contentsname}{\centering Table of Contents of Appendix}
\tableofcontents

\setlength{\tabcolsep}{8pt}

\vskip 2em
\section{Additional Related Work} \label{appendix:related-work}

\paragraph{Applications of PRMs.}
PRMs can be used to supervise LLM reasoning during training and inference. For {\bf training}, PRMs can generate reward signals, particularly in reinforcement learning settings~\citep{pan2023lets, zhang2024restmcts}. They can be applied either to re-rank candidate responses from the policy or to provide direct reward~\citep{uesato2022solving}. For {\bf inference}, PRMs can guide response selection and refinement through Best-of-K~\citep{li-etal-2023-making}, self-correction~\citep{saunders2022selfcritiquing, madaan2023selfrefine}, and step-level search~\citep{ma2023letsrewardstep, snell2024scaling}.

\paragraph{Outcome-based training for PRMs.}
Provided the difficulty of collecting step-level error labels, recent work proposes methods to train PRMs without using step-level labels. \citet{yuan2025free} proposes implicit PRMs that can be trained only using final answers. They theoretically show that ORMs trained with the reward that is parameterized by the log-likelihood ratio of two causal language models~(e.g.,~DPO~\citep{rafailov2023direct}) implicitly learns a Q function and can be used as PRMs. In addition, recent work~\citep{feng2025prmnecessary, kim2025scaling} reports that existing Large Reasoning Models like DeepSeek-R1~\citep{deepseekai2025deepseekr1} have strong process-level rewarding capabilities on mathematical reasoning tasks, while they are not explicitly trained for process-level rewarding.

\paragraph{Formal logic and theorem proving.}
Formal logic tasks have been used to evaluate and improve the reasoning of LLMs~\citep{clark2021ruletakers, tafjord-etal-2021-proofwriter, morishita2024enhancing}. These tasks are often formulated in logical languages compatible with automated solvers such as Z3~\citep{demoura2008z3}, Vampire~\citep{kovacs2013vampire}, or E~\citep{schulz2002e}. Separately, formal theorem proving has been studied in the context of LLM-based proof generation~\citep{polu2020generative, kaiyu2023leandojo, xin2024deepseekprover}, where proof assistants such as Isabelle/HOL~\citep{nipkow2002isabelle}, Coq~\citep{Coq-refman}, and Lean~\citep{lean4} are used to validate generated proofs. In contrast, our work presents the first attempt to use verification outputs from these tools as training signals for reward models. Moreover, while these tools are designed and have been used for solution-level verification, we introduce a novel use of them to provide step-level supervision for PRMs.

\section{Examples from \fover-40K} \label{appendix:fover-dataset-example}

We provide examples from \fover-40K.

\paragraph{Formal logic.}
Here is an example of the data in the formal logic task based on the initial responses from Llama~3.1~8B.

\vskip 1em
\begin{minted}[frame=single, breaklines=true, fontsize=\scriptsize, linenos=false]{text}
{
  'id': 'fldx2-train-020948_Llama-3.1-8B-Instruct',
  'error_labels': [True, False],
  'messages': [
  {'role': 'user', 'content': '** Problem **
    Based on the provided facts ($context$), either prove or disprove the hypothesis or state that it is unknown. The facts and the hypothesis are written in logical formulas as follows: capital letters such as "{A}", "{B}", "{AB}" are predicates, small letters such as "{a}", "{b}", "{ab}" are constants, "&" is logical conjunction, "v" is logical disjunction, "¬" is negation, "->" is implication, "(x)" is "for all x", and "(Ex)" is "for some x".
    
    $hypothesis$: ¬{B}
    
    $context$:\nfact1: ¬(¬{F} & ¬{G}) -> ¬{E}
    fact2: ¬{CF}\nfact3: ¬{IG}
    fact4: ¬{DQ}
    fact5: ¬{CN}\nfact6: ¬{BA}
    fact7: {B} -> (¬{DC} & ¬{A})
    fact8: ¬{DM}\nfact9: {C} -> {B}
    fact10: ¬{IF}
    fact11: ¬{JJ}
    fact12: (¬{GL} & ¬{AP})
    fact13: ¬{E} -> ({D} & {C})
    fact14: (¬{A} & ¬{B})
    fact15: (¬{DC} & ¬{L})
    fact16: ¬{IJ}
    fact17: ¬{EM}
    fact18: {H} -> ¬(¬{F} & ¬{G})
    fact19: ¬{DO}
    
    ** Task **
    Your task is to evaluate the accuracy of each step in the provided solution to the above question. For each step, respond with "correct" if the reasoning is logically valid and mathematically sound, or if the step is a general statement or transition that does not contain reasoning. Respond with "incorrect" if the step includes any errors or flawed logic.
    
    ** Sotluion **
    fact7 -> int1: {B} -> (¬{DC} & ¬{A})'},
    {'role': 'assistant', 'content': 'correct'},
    {'role': 'user', 'content': 'fact6 -> int2: ¬{B}'},
    {'role': 'assistant', 'content': 'incorrect'}
  ],
}
\end{minted}

\paragraph{Formal theorem proving.}
Here is an example of the data in the formal proof task based on the initial responses from Qwen~2.5~7B.

\vskip 1em
\begin{minted}[frame=single, breaklines=true, fontsize=\scriptsize, linenos=false]{text}
{
  'id': 'metamathqa_GSM_Rephrased_076554_Qwen2.5-7B-Instruct',
  'error_labels': [True, True, True, True, True, True, False],
  'messages': [
    {'role': 'user', 'content': '** Problem **
    Generate a proof for the following theorem in the Isabelle proof assistant format.
    
    theorem example:
        assumes "(Jorge_age_2005::nat) = 16"
        and "(Jorge_age_difference::nat) = 24"
        and "(Jorge_current_age::nat) = Jorge_age_2005 + (2022 - 2005)"
        and "(Simon_current_age::nat) = Jorge_current_age + Jorge_age_difference"
        and "(Simon_age_2010::nat) = Simon_current_age + (2010 - 2022 + 2005)"
        shows "Simon_age_2010 = 50"
        
    ** Task **\nYour task is to evaluate the accuracy of each step in the provided solution to the above question. For each step, respond with "correct" if the reasoning is logically valid and mathematically sound, or if the step is a general statement or transition that does not contain reasoning. Respond with "incorrect" if the step includes any errors or flawed logic.
    
    ** Sotluion **
    We\'ll start from the beginning and proceed logically.'},
    {'role': 'assistant', 'content': 'correct'},
    {'role': 'user', 'content': 'have "Jorge_current_age = 16 + (2022 - 2005)" using assms by simp'},
    {'role': 'assistant', 'content': 'correct'},
    {'role': 'user', 'content': 'then have "Jorge_current_age = 16 + 17" using assms by simp\nthen have "Jorge_current_age = 33" using assms by simp'},
    {'role': 'assistant', 'content': 'correct'},
    {'role': 'user', 'content': 'then have "Simon_current_age = 33 + 24" using assms by simp'},
    {'role': 'assistant', 'content': 'correct'},
    {'role': 'user', 'content': 'then have "Simon_current_age = 57" using assms by simp'},
    {'role': 'assistant', 'content': 'correct'},
    {'role': 'user', 'content': 'then have "Simon_age_2010 = 57 + (2010 - 2022 + 2005)" using assms by simp'},
    {'role': 'assistant', 'content': 'correct'},
    {'role': 'user', 'content': 'then have "Simon_age_2010 = 57 + 83" using assms by simp'},
    {'role': 'assistant', 'content': 'incorrect'}
  ],
}
\end{minted}
\vskip 1em

\begin{figure}[t]
\begin{subfigure}{\linewidth}
    \centering
    \includegraphics[width=.6\linewidth]{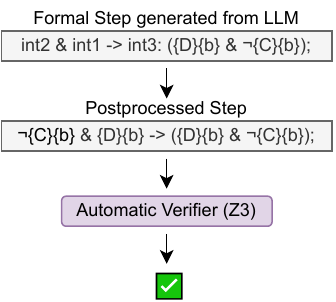}
    \caption{Formal logic task.}
    \label{fig:z3-verification}
\end{subfigure}
\vskip .5em
\begin{subfigure}{\linewidth}
    \centering
    \includegraphics[width=.6\linewidth]{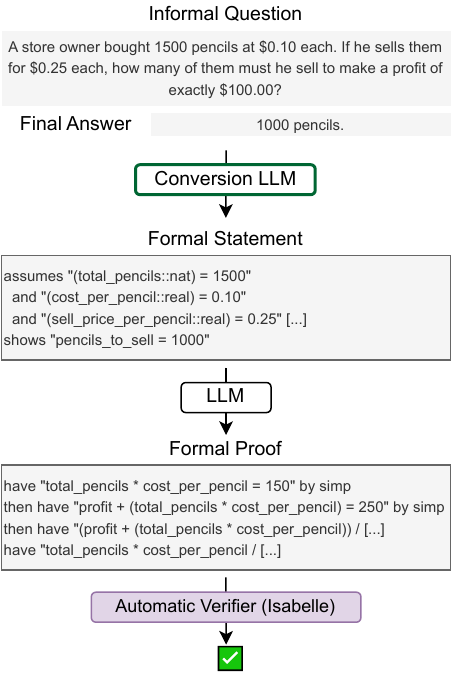}
    \caption{Formal theorem proving task.}
    \label{fig:isabelle-conversion}
\end{subfigure}
\caption{Automatic step-level error annotation for \fover-40K by formal verification tools.}
\end{figure}

\section{Creation Process of \fover-40K} \label{appendix:dataset-creation-fover40k}

This section provides details of the creation process of \fover-40K, which is outlined in Section~\ref{sec:fover-method}.

\subsection{Formal Logic}  \label{appendix:dataset-creation-formal-logic}

\paragraph{Formal solution generation~(Appendix~\ref{appendix:fldx2-initial-generation}).}
We prompt Llama~3.1~8B and Qwen~2.5~7B to generate step-by-step formal solutions to FLDx2 in a format compatible with Z3. We use a few-shot instruction to guide LLMs to follow the format and filter out syntactically invalid solutions using Z3.

\paragraph{Automatic error annotation~(Appendix~\ref{appendix:fldx2-verification}).}
Z3 is designed for verification at the solution level, but we use it at the step level by supplying Z3 with the premises and conclusion for the target step to determine logical validity, as in Figure~\ref{fig:z3-verification}.

\subsubsection{Initial Response Generation} \label{appendix:fldx2-initial-generation}

First, we need to generate symbolic solutions from LLMs, which may include logical mistakes, but they should be in a valid format compatible with Z3~\citep{demoura2008z3}.

We use FLDx2~\citep{morishita2023learning, morishita2024enhancing} as a base dataset for our formal logic task. We use the symbolic version of the dataset. To simplify the verification pipeline, we removed cases whose reasoning steps include ``assump,'' which is used in cases such as proof by contradiction.

We generate formal solutions from Llama~3.1~8B and Qwen~2.5~7B. The following is an example of a few-shot demonstration for the initial generation. We provide six examples as a demonstration. After generating the formal solutions, we filter out those with an invalid format using Z3.

\begin{minted}[frame=single, breaklines=true, fontsize=\scriptsize, linenos=false]{text}
{
  'role': 'user',
  'content':
    'Based on the provided facts ($context$), either prove or disprove the hypothesis or state that it is unknown. The facts and the hypothesis are written in logical formulas as follows: capital letters such as "{A}", "{B}", "{AB}" are predicates, small letters such as "{a}", "{b}", "{ab}" are constants, "&" is logical conjunction, "v" is logical disjunction, "¬" is negation, "->" is implication, "(x)" is "for all x", and "(Ex)" is "for some x".
    
    Don't generate anything other than the proof and proof_label.
    Don't generate the proof by contradiction. If the hypothesis is disproved, provide a direct counterexample. If the proof_label is PROVED, your proof should end with the 'hypothesis'. If the proof_label is DISPROVED, your proof should end with '¬hypothesis'.
    
    $hypothesis$: ({AB}{c} & ¬{B}{c})
    
    $context$:
    fact1: (x): ¬{D}x -> ({A}x & {C}x)
    fact2: ¬({B}{a} & ¬{A}{a})
    fact3: ¬{AB}{b} -> ¬{C}{d}
    fact4: ¬{C}{a} -> ¬{A}{d}
    fact5: ¬{A}{a}
    fact6: ({AB}{c} & ¬{B}{c}) -> {C}{d}
    fact7: ¬({F}{b} v {D}{b}) -> ¬{D}{a}
    fact8: {A}{a} -> ({AB}{c} & ¬{B}{c})
    fact9: ({AB}{c} & ¬{AA}{c}) -> {B}{b}
    fact10: ({AB}{c} & ¬{B}{c}) -> {C}{b}'
},
{
  'role': 'assistant',
  'content':
    '$proof$:
    fact6 -> int1: ¬{C}{d} -> ¬({AB}{c} & ¬{B}{c});
    fact3 & int1 -> int2: ¬{AB}{b} -> ¬({AB}{c} & ¬{B}{c});

    $proof_label$: UNKNOWN'
}
\end{minted}

\subsubsection{Automatic Step-level Annotation} \label{appendix:fldx2-verification}

Second, we automatically annotate step-level error labels, which will be used to train PRMs. We use Z3~\citep{demoura2008z3} for annotating step-level error labels. Specifically, we use the checker provided by the FLDx2 paper~\citep{morishita2024enhancing} based on Z3.\footnote{\url{https://github.com/hitachi-nlp/FLD-generator/blob/00d12c4a9132a4fb43cd77f24db03ea7f5b27877/FLD\_generator/formula\_checkers/z3\_logic\_checkers/checkers.py\#L179}} Z3 is originally designed for solution-level verification, so we need to write code to use it for step-level verification. As explained in Figure~\ref{fig:z3-verification}, we first postprocess each step in solutions to an independent logical step and check the validity using Z3.

\subsection{Formal Theorem Proving} \label{appendix:dataset-creation-formal-proof}

\paragraph{Formal statement generation~(Appendix~\ref{appendix:math-proof-dataset}).} We first generate formal statements in the format compatible with Isabelle from GSM8K-level problems, including GSM8K~\citep{cobbe2021gsm8k}, GSM8K-based cases in MetaMathQA~\citep{yu2024metamath}, and math word problems in Big-Math~\citep{albalak2025bigmath}, using Qwen~2.5~7B.

\paragraph{Formal solution generation~(Appendix~\ref{appendix:formal-proof-generation}).} We instruct Qwen~2.5~7B to generate step-by-step formal proofs in the format compatible with Isabelle to the formal statements.

\paragraph{Automatic error annotation~(Appendix~\ref{appendix:isabelle-step-level}).}
Isabelle is designed for solution-level verification. To obtain step-level error labels, we implement wrapper code for step-level verification. Our code assumes that the other steps are correct when evaluating the target step.

\subsubsection{Formal Statement Generation} \label{appendix:math-proof-dataset}

First, we generate formal statements in the format compatible with Isabelle~\citep{nipkow2002isabelle}. Motivated by prior work~\citep{zhou2024dont}, we generate formal statements from math word problems using Qwen~2.5~7B with few-shot prompting. Specifically, we convert informal GSM8K-level math word problems: GSM8K~\citep{cobbe2021gsm8k}, GSM8K-based cases in MetaMathQA~\citep{yu2024metamath}, and math word problems in Big-Math~\citep{albalak2025bigmath}, into formal statements.

\subsubsection{Formal Proof Generation} \label{appendix:formal-proof-generation}

We provide an instruction with few-shot examples to generate formal proofs from Qwen~2.5~7B. The following is an example of a few-shot demonstration for the formal proof generation.

\begin{minted}[frame=single, breaklines=true, fontsize=\scriptsize, linenos=false]{text}
{
  'role': 'user',
  'content': '** Formal Statement **
theorem example:
    assumes "(Cabin_cost::nat) = 129000"
        and "(Gloria_cash::nat) = 150"
        and "(Cypress_count::nat) = 20"
        and "(Pine_count::nat) = 600"
        and "(Maple_count::nat) = 24"
        and "(Cypress_price::nat) = 100"
        and "(Maple_price::nat) = 300"
        and "(Pine_price::nat) = 200"
        and "(After_paying_Alfonso::nat) = Gloria_cash + Cypress_count * Cypress_price + Maple_count * Maple_price + Pine_count * Pine_price - Cabin_cost"
    shows "After_paying_Alfonso = 200"'
},
{
  'role': 'assistant',
  'content': 'proof -
    have "Cypress_count * Cypress_price = 2000"
        using assms by simp
    then have "Maple_count * Maple_price = 7200"
        using assms by simp
    then have "Pine_count * Pine_price = 120000"
        using assms by simp
    then have "Gloria_cash + Cypress_count * Cypress_price + Maple_count * Maple_price + Pine_count * Pine_price = 129200"
        using assms by simp
    then have "Gloria_cash + Cypress_count * Cypress_price + Maple_count * Maple_price + Pine_count * Pine_price - Cabin_cost = 200"
        using assms by simp
    then have "After_paying_Alfonso = 200"
        using assms by simp
    show ?thesis
        using assms by simp
qed'
}
\end{minted}

\subsubsection{Automatic Step-level Proof Verification} \label{appendix:isabelle-step-level}

\paragraph{Format verification.}
As we use LLMs for the conversion, it is possible that the conversion generates proofs in an invalid format. To filter out theorems in the invalid format, we use the ``sorry'' keyword of Isabelle. The ``sorry'' keyword serves as a placeholder for incomplete or unproven proofs, allowing the theorem to be accepted by the system without a formal justification. By inserting ``sorry'' into all generated proof steps, we can isolate and verify only the syntactic correctness theorems.

For example, in the following proof, \texttt{babysitting\_minutes × (Weng\_hourly\_wage / 60)} contains the symbol \texttt{×}, which is not a valid multiplication operator in Isabelle syntax.

\begin{minted}[frame=single, breaklines=true, fontsize=\scriptsize, linenos=false, highlightlines={10}, highlightcolor=Thistle]{text}
theorem example:
    assumes "(Weng_hourly_wage::real) = 12"
        and "(babysitting_minutes::real) = 50"
        and "(babysitting_hours::real) = babysitting_minutes / 60"
        and "(Weng_earnings::real) = Weng_hourly_wage * babysitting_hours"
    shows "Weng_earnings = 10"
proof -
    have "Weng_hourly_wage / 60 = 0.20"
        sorry
    then have "babysitting_minutes × (Weng_hourly_wage / 60) = 10"
        sorry
    then have "Weng_earnings = 10"
        sorry
    thus ?thesis
        sorry
qed
\end{minted}

\noindent
For this input, Isabelle returns the following error.

\begin{minted}[frame=single, breaklines=true, fontsize=\scriptsize, linenos=false, highlightlines={10}, highlightcolor=Thistle]{text}
Step error: Inner syntax error (line 1)\nat \"? ( Weng_hourly_wage / 60 ) = 10\"\nFailed to parse prop\nAt command \"have\" (line 1)
\end{minted}

\paragraph{Step-level verification.}
By default, Isabelle halts at the first encountered error and does not provide a step-by-step verification of a proof. To enable independent verification of each step in a multi-step proof, we insert the ``sorry'' keyword in all but one step. This allows Isabelle to type-check and parse each step individually, even if other steps are incomplete or invalid.

The following example is for verifying the third step independently. For each theorem, we run Isabelle once per step.

\begin{minted}[frame=single, breaklines=true, fontsize=\scriptsize, linenos=false, highlightlines={15}, highlightcolor=Thistle]{text}
theorem example:
    assumes "(wallet_cost::nat) = 100"
        and "(betty_savings::nat) = wallet_cost div 2"
        and "(parent_contribution::nat) = 15"
        and "(grandparent_contribution::nat) = 2 * parent_contribution"
        and "(total_savings::nat) = betty_savings + parent_contribution + grandparent_contribution"
        and "(additional_needed::nat) = wallet_cost - total_savings"
    shows "additional_needed = 5"
proof -
    have "betty_savings = wallet_cost div 2"
        sorry
    then have "betty_savings = 50"
        sorry
    have "grandparent_contribution = 2 * parent_contribution"
        by simp
    then have "grandparent_contribution = 30"
        sorry
    then have "parent_contribution + grandparent_contribution = 45"
        sorry
    then have "total_savings = 95"
        sorry
    then have "additional_needed = wallet_cost - total_savings"
        sorry
    then have "additional_needed = 5"
        sorry
    thus ?thesis
        sorry
qed\end{minted}

\newcommand{\validationdatalistshort}{Orca-Math~\citep{mitra2024orcamath} and two tasks in BBH~\citep{suzgun-etal-2023-challenging}}
\newcommand{\validationdatalist}{\validationdatalistshort~(Logical Deduction~(three objects) and Boolean Expressions)}

\section{\fover{} Beyond \fover-40K} \label{appendix:extension}

As the first work towards this direction, we evaluate \fover{} by creating PRM training data using relatively simple tasks with a minimal pipeline to keep the evaluation focused and clear. However, \fover{} is not limited to the tasks and tools we used in this paper and can be extended to create training data using different types of tasks or more complex tasks. When applying \fover{} to create PRM training data using more complex tasks, there are two potential challenges: (1)~generating formal solutions from LLMs in a valid format compatible with formal verification tools and (2)~verifying step-level correctness using formal verification tools.

First, formal solutions should be in a syntactically valid format compatible with formal verification tools we use for verification. Following syntactical rules and formats of some formal verification tools can be challenging for LLMs, especially when we target more complex problems. Recent LLMs are increasingly capable of generating formal solutions in valid formats, showing strong performance in first-order logic~\citep{olausson-etal-2023-linc} and a growing ability to produce syntactically valid formal proofs~\citep{ren2025deepseekproverv2}. We expect future models to further improve their capabilities to generate formal solutions and be more suitable for creating PRM training data using \fover.

Second, we need to make formal verification tools verify step-level correctness. The tools are often designed for solution-level verification, so we often need to adapt them for step-level verification, as we did in this paper for creating \fover-40K. When creating PRM training data using more complex tasks, we may need to further modify the verification pipeline to support new operations.

For example, to keep the verification pipeline simple, we did not use problems that involve assumptions in the formal logic task~(FLDx2), such as proofs by contradiction, when creating \fover-40K. However, we can extend our verification pipeline to support such cases. Existing verification tools are already capable of performing solution-level verification for proofs by contradiction, so we can make use of them to provide step-level verification. When handling assumptions in our framework, the type of mistake that cannot be detected through our current step-independent verification alone is illustrated by the following example, because the step-independent verification assumes that preceding intermediate results are correct:

\begin{itemize}
    \item Premises: fact1: B; fact2: B$\rightarrow$C; fact3: C$\rightarrow$A;
    \item Hypothesis: A
    \item fact1 and fact2 $\rightarrow$ C; \textbf{Assume A; A $\rightarrow$ Hypothesis;} Therefore, the hypothesis is proved.
\end{itemize}

\noindent
In this case, the existing solution-level verification will identify this solution as an error because the assumption is not properly discharged. Thus, by combining step-independent verification with solution-level verification, we can identify and label the final step as erroneous.

\section{Implementation Details}

This section provides details of our experiments.

\subsection{Input Format and Postprocessing} \label{appendix:fover-prm}

\paragraph{\fover-PRMs and the baseline PRMs.}
We create inputs to LLM-based PRMs by preprocessing step-by-step solutions into a conversation format where each input contains a single step, and the expected output is a single token: ``correct'' or ``incorrect''. To obtain step-level scores, we extract logits for the two words and apply the softmax function to compute the prediction probability for ``correct''. This approach follows prior work~\citep{xiong2024rlhflowmath} that uses LLMs as PRMs. As the baseline PRMs are not fine-tuned, we provide zero-shot instructions about this format.

First, we describe the input format for \fover-PRMs and the baseline LLM-based PRMs, which are based on Llama~3.1~8B and Qwen~2.5~7B. \fover-PRMs are trained on \fover-40K, so the input format has the same format as the training data. The only difference is that we replace all step-level labels with ``correct'' in the input. This preprocessing allows us to provide the whole input once to get the step-level predictions for all steps.
The following is an example input for GSM8K.

\begin{minted}[frame=single, breaklines=true, fontsize=\scriptsize, linenos=false, highlightlines={13,15,17,19,21}, highlightcolor=Thistle]{text}
[
  {
    'role': 'user',
    'content': '** Problem **
      Alice is 7 years older than Beth, who is 5 years older than Erica. What is the difference between the ages of Alice and Erica, if Erica is 30 years old?

      ** Task **
      Your task is to evaluate the accuracy of each step in the provided solution to the above question. For each step, respond with "correct" if the reasoning is logically valid and mathematically sound, or if the step is a general statement or transition that does not contain reasoning. Respond with "incorrect" if the step includes any errors or flawed logic.

      ** Sotluion **
      Since Erica is 30 years old, and Beth is 5 years older than Erica, Beth is 30 + 5 = 35 years old.'
  },
  {'role': 'assistant', 'content': 'correct'},
  {'role': 'user', 'content': 'Alice is 7 years older than Beth, who is 35 years old. '},
  {'role': 'assistant', 'content': 'correct'},
  {'role': 'user', 'content': 'So, Alice is 35 + 7 = 42 years old.'},
  {'role': 'assistant', 'content': 'correct'},
  {'role': 'user', 'content': "The difference between Alice's age and Erica's age is 42 - 30 = 12 years."},
  {'role': 'assistant', 'content': 'correct'},
  {'role': 'user', 'content': 'Therefore, the answer (arabic numerals) is 12.'},
  {'role': 'assistant', 'content': 'correct'}
]
\end{minted}

Next, we describe the postprocessing for \fover-PRMs and the baseline LLM-based PRMs.

\paragraph{Extracting logits.}
Since we use causal LLMs as PRMs, we extract the model’s predictions for the tokens immediately preceding the dummy step-level labels (e.g., ``correct'') in the input.

\paragraph{Computing step-level scores.}
At each identified position, we extract the logits corresponding to the tokens ``correct'' and ``incorrect''. We then apply the softmax function over these two logits to compute the probability assigned to ``correct''. This probability serves as the step-level score.

\paragraph{PRMs based on Llama~3.1~8B.}
In RLHFlow-Llama3.1-8B-DeepSeek and RLHFlow-Llama3.1-8B-Mistral~\citep{xiong2024rlhflowmath}, the input format is mostly similar to ours, with the key difference being the use of ``+'' and ``-'' instead of ``correct'' and ``incorrect''.\footnote{\url{https://github.com/RLHFlow/RLHF-Reward-Modeling/tree/main/math-rm}} For these models, we apply our input format and postprocessing procedures with a simple substitution of ``correct'' with ``+''.

\paragraph{PRMs based on Qwen~2.5~7B.}
Qwen2.5-Math-7B-PRM800K~\citep{zheng2024processbench} and Qwen2.5-Math-PRM-7B~\citep{zhang2025lessons} are supported by vLLM~\citep{kwon2023efficient}. We follow the input format specified in their respective model descriptions and adopt the reward modeling in vLLM.\footnote{\url{https://docs.vllm.ai/en/latest/models/pooling_models.html}} For Qwen2.5-7B-Skywork-PRM~\citep{skyworkopeno12024}, we use a code provided by the authors.\footnote{\url{https://github.com/SkyworkAI/skywork-o1-prm-inference}}

\subsection{Training Settings} \label{appendix:training}

This section provides details of the training settings for our \fover-PRMs.

\paragraph{Balanced labels.}
The last steps of the solutions in \fover-40K includes balanced step-level error labels of 50\% of ``correct'' and ``incorrect'' labels. Then, we can perform training on the balanced label by setting \texttt{mask\_history:~True} in LLaMA-Factory, which configures the model to use only the last step of each conversation during training.

\paragraph{Training parameters.}
We fine-tune all model parameters and do not use parameter-efficient techniques. We use the AdamW optimizer~\citep{loshchilov2018decoupled} and select the learning rate based on the average Best-of-K performance on the validation tasks: \validationdatalist. We evaluate models trained with the learning rate 1e-6, 2e-6, 5e-6, and 1e-5, and select the model with the best average performance on the validation tasks.
We use the parameters in Table~\ref{tab:hyperparameters} in all models, and we did not conduct hyperparameter tuning for these parameters. Please refer to the configuration files in our code for further details.

\begin{table}[t]
    \centering
    \small
    \begin{tabular}{ll}
    \toprule
        Parameter                                & Value  \\
    \midrule
        Number of Epochs                         & 1      \\
        Batch size                               & 32     \\
        Learning Rate Warm up and Decay Strategy & Linear \\
        Learning Rate Warm up Ratio              & 0.5    \\
    \bottomrule
    \end{tabular}
\caption{Hyperparameters for training on \fover-40K}
\label{tab:hyperparameters}
\end{table}

\begin{table*}[t]
    \centering
    \scriptsize
    \begin{tabular}{cc}
    \toprule
        Dataset & Source \\
    \midrule
        GSM8K            & \url{https://huggingface.co/datasets/openai/gsm8k} \\
        MATH             & \url{https://github.com/hendrycks/math}  \\
        AQuA-RAT         & \url{https://huggingface.co/datasets/deepmind/aqua_rat} \\
        AIME (2016-2024) & \url{https://huggingface.co/datasets/di-zhang-fdu/AIME\_1983\_2024} \\
    \midrule
        FOLIO            & \url{https://huggingface.co/datasets/yale-nlp/FOLIO} \\
        LogicNLI         & \url{https://huggingface.co/datasets/tasksource/LogicNLI} \\
    \midrule
        ANLI             & \url{https://huggingface.co/datasets/facebook/anli} \\
        HANS             & \url{https://github.com/tommccoy1/hans} \\
    \midrule
        MMLU             & \url{https://huggingface.co/datasets/sam-paech/mmlu-pro-nomath-sml} \\
    \midrule
        BBH              & \url{https://github.com/suzgunmirac/BIG-Bench-Hard/tree/main/bbh} \\
    \bottomrule
    \end{tabular}
\caption{Reasoning benchmarks evaluated in the Best-of-K experiments.}
\label{tab:evaluation-datasets}
\end{table*}

\begin{table*}[t!]
    \centering
    \scriptsize
    \begin{tabular}{cM{.3\linewidth}M{.45\linewidth}}
    \toprule
        Dataset  & Few-shot Examples for Initial Generation & Answer Matching \\
    \midrule
        GSM8K    & \citet{kojima2022large} \tablefootnote{\url{https://github.com/kojima-takeshi188/zero\_shot\_cot/blob/5ef330fcdeec0cd26aee27943504f91f8ec1c33c/utils.py\#L328}} &
        Exact match after extraction and conversion to integer \\
        MATH     & \citet[][Appendix~D.2]{lewkowycz2022solving} & \citet[][Appendix~G]{lewkowycz2022solving} \\
        AQuA-RAT & Made by us (3-shot) & Exact match after extraction \\
        AIME     & Made by us (3-shot) & Exact match after extraction and conversion to integer \\
    \midrule
        FOLIO    & Made by us (2-shot) & Exact match after extraction \\
        LogicNLI & Made by us (3-shot) & Exact match after extraction \\
    \midrule
        ANLI     & Made by us (3-shot) & Exact match after extraction \\
        HANS     & Made by us (2-shot) & Exact match after extraction \\
    \midrule
        MMLU     & Made by us (4-shot) & Exact match after extraction \\
    \midrule
        BBH      & \citet{suzgun-etal-2023-challenging}\tablefootnote{\url{https://github.com/suzgunmirac/BIG-Bench-Hard/blob/main/cot-prompts}} & Exact match after extraction \\
    \bottomrule
    \end{tabular}
\caption{Detailed settings for Best-of-K downstream evaluation}
\label{tab:downstream-evaluation-datasets}
\end{table*}

\subsection{Evaluation Settings of Best-of-K}

\paragraph{Benchmarks.}
Table~\ref{tab:evaluation-datasets} shows sources of reasoning benchmarks evaluated in Best-of-K in Section~\ref{sec:experiments}.

\paragraph{Initial generation prompts.}
Table~\ref{tab:downstream-evaluation-datasets} shows detailed settings of generating $K=7$ responses for the Best-of-K evaluation in Section~\ref{sec:experiments}. We create new few-shot examples or modify few-shot demonstrations used in prior work to enhance the quality and to simplify the post-processing procedure. For example, we add line breaks between reasoning steps in all tasks. An example prompt for GSM8K is provided in Appendix~\ref{appendix:fldx2-initial-generation}. Please also refer to our code for further details.

\newcommand{\sampleandrankheaderfirst}{
\multicolumn{4}{c}{Math} & \multicolumn{2}{c}{Logic} & \multicolumn{2}{c}{NLI} & \multicolumn{1}{c}{MMLU} & \multicolumn{3}{c}{BBH} & \multirow{2.7}{*}{Average}
}
\newcommand{\sampleandrankheadermidrule}[1]{
#1 \cmidrule(l{2pt}r{2pt}){3-6} \cmidrule(l{2pt}r{2pt}){7-8} \cmidrule(l{2pt}r{2pt}){9-10} \cmidrule(l{2pt}r{2pt}){11-11} \cmidrule(l{2pt}r{2pt}){12-14}
}
\newcommand{\sampleandrankheadersecond}{
GSM8K & MATH & AQuA & AIME & FOLIO & LogicNLI & ANLI & HANS & {\fontsize{5pt}{5pt}\selectfont Pro-NoMath} & Temporal & Tracking & Sorting
}

\begin{table*}[t!]
    \centering
    \scriptsize
    \setlength{\tabcolsep}{1.8pt}
    \begin{tabular}{cccM{.2\linewidth}}
    \toprule
        PRMs & Source & Base Datasets & Training Data Creation \\
    \midrule
        RLHFlow-Llama3.1-8B-DeepSeek~\citeyearpar{xiong2024rlhflowmath}  & \href{https://huggingface.co/RLHFlow/Llama3.1-8B-PRM-Deepseek-Data}{RLHFlow/Llama3.1-8B-PRM-Deepseek-Data} & GSM8K, MATH & Math-Shepherd~\citeyearpar{wang-etal-2024-math-shepherd} \\
    \midrule
        RLHFlow-Llama3.1-8B-Mistral~\citeyearpar{xiong2024rlhflowmath}  & \href{https://huggingface.co/RLHFlow/Llama3.1-8B-PRM-Mistral-Data}{RLHFlow/Llama3.1-8B-PRM-Mistral-Data} & GSM8K, MATH & Math-Shepherd~\citeyearpar{wang-etal-2024-math-shepherd} \\
    \midrule
        Qwen2.5-Math-7B-PRM800K~\citeyearpar{zheng2024processbench} & \href{https://huggingface.co/Qwen/Qwen2.5-Math-7B-PRM800K}{Qwen/Qwen2.5-Math-7B-PRM800K} & MATH & Human annotation \\
    \midrule
        Qwen2.5-Math-PRM-7B~\citeyearpar{zhang2025lessons} & \href{https://huggingface.co/Qwen/Qwen2.5-Math-PRM-7B}{Qwen/Qwen2.5-Math-PRM-7B} & Private Data & Math-Shepherd~\citeyearpar{wang-etal-2024-math-shepherd} \& LLM-as-a-Judge \\
    \midrule
        Qwen2.5-7B-Skywork-PRM~\citeyearpar{skyworkopeno12024} &
        \href{https://huggingface.co/Skywork/Skywork-o1-Open-PRM-Qwen-2.5-7B}{Skywork/Skywork-o1-Open-PRM-Qwen-2.5-7B} &
        Hidden & Hidden
        \\
    \bottomrule
    \end{tabular}
    \caption{PRMs we evaluate in Section~\ref{sec:experiments}.}
    \label{tab:sota-prm}
\end{table*}

\subsection{Computational Resources} \label{appendix:computational-resources}

We use four NVIDIA A100 SXM4 80GB GPUs for training and inference. Training each 8B-class model on our dataset takes approximately one hour using our training data, \fover-40K. Evaluation takes considerably more time because the Best-of-K evaluation requires multiple candidate solutions, and reproducing all the evaluation in this paper will take approximately three days.

Verification using Z3 and Isabelle does not require GPUs and can be run in parallel on CPU-only servers, enabling efficient and scalable PRM training data creation. On average, in a single process, Z3 verifies all datasets in less than 10 minutes for the formal logic task, and Isabelle verifies each step in approximately 3 seconds for the formal theorem proving tasks. For annotation, we run one Z3 process for formal logic and 40 parallel Isabelle processes for formal theorem proving on AMD EPYC 7763 64-core processors.

\subsection{Model Access and Software Libraries} \label{appendix:model-access}

\paragraph{LLMs.}
We use LLMs from Hugging~Face~Hub. We use \texttt{meta-llama/Llama-3.1-8B-Instruct} and \texttt{Qwen/Qwen2.5-7B-Instruct} as base models for our PRMs. We also use these models to generate initial responses used in creating \fover-40K, and also for generating $K=7$ responses in the Best-of-K evaluation~(\S\ref{sec:experiments}).

\paragraph{Existing PRMs.}
Details for the PRMs evaluated in Section~\ref{sec:experiments} are listed in Table~\ref{tab:sota-prm}. We acquire these models at Hugging~Face~Hub and use vLLM~\citep{kwon2023efficient} to generate reward scores.\footnote{\url{https://docs.vllm.ai/en/latest/models/supported_models.html\#reward-modeling-task-reward}}

\paragraph{Software libraries.}
For inference, we use vLLM~\citep{kwon2023efficient} for accelerating LLM inference. 
For training, we use LLaMA-Factory~\citep{zheng2024llamafactory} for fine-tuning PRMs.

\lstset{
  basicstyle=\ttfamily\small,
  breaklines=true,
  breakatwhitespace=false,
  frame=single,
  columns=fullflexible
}

\section{Ethical Considerations}
\paragraph{Social impacts.}
This research does not involve human subjects or private data. This paper proposes a method for improving PRMs, which detect mistakes in LLM responses. We do not expect any harmful impact.

\paragraph{Dataset.}
Our dataset, \fover-40K, is created from formal verification tools and logic and mathematical reasoning datasets. We do not expect that our dataset includes information that uniquely identifies individual people or offensive content.

\paragraph{The use of Large Language Models.}
In paper writing, we used LLMs to polish writing. We used ChatGPT-5 via OpenAI's web interface.

\section{Reproducibility Statement and License}

\ifreview
\else
The datasets, models, and code are provided at~\url{https://github.com/psunlpgroup/FoVer}.
\fi

\ifreview
\else
We release our dataset under Creative Commons Attribution 4.0 International and our code under Apache License 2.0. Our dataset and code are based on the following resources. We consider our license to be (one-way) compatible with all licenses listed below.
\fi

\paragraph{Datasets.}
\fover-40K is based on the following datasets.

\begin{itemize}
    \item FLDx2~\citeyearpar{morishita2024enhancing}: CC BY 4.0\footnote{\url{https://github.com/hitachi-nlp/FLD-corpus/blob/neurips_2025/LICENSE}}
    \item GSM8K~\citeyearpar{cobbe2021gsm8k}: MIT\footnote{\url{https://github.com/openai/grade-school-math/blob/master/LICENSE}}
    \item MetaMathQA~\citeyearpar{yu2024metamath}: MIT\footnote{\url{https://huggingface.co/datasets/meta-math/MetaMathQA/blob/main/README.md}}
    \item Big-Math~\citeyearpar{albalak2025bigmath}: Apache License 2.0\footnote{\url{https://huggingface.co/datasets/SynthLabsAI/Big-Math-RL-Verified/blob/main/README.md}}
\end{itemize}

\paragraph{Code and packages.}
Our code is partially based on the following resources.

\begin{itemize}
    \item FLD~\citeyearpar{morishita2024enhancing}: Apache License 2.0\footnote{\url{https://github.com/hitachi-nlp/FLD/blob/neurips_2025/LICENSE}}
    \item Isabelle: BSD-style regulations\footnote{\url{https://isabelle.in.tum.de/}}
    \item Neural theorem proving tutorial~\citeyearpar{ntptutorial}: MIT\footnote{\url{https://github.com/wellecks/ntptutorial/blob/main/LICENSE}}
    \item DTV~\citeyearpar{zhou2024dont}: MIT\footnote{\url{https://github.com/jinpz/dtv/blob/main/LICENSE}}
\end{itemize}

\end{document}